\pdfoutput=1

\documentclass[11pt]{article}

\usepackage[utf8]{inputenc} % allow utf-8 input
\usepackage[T1]{fontenc}    % use 8-bit T1 fonts
\usepackage{hyperref}       % hyperlinks
\usepackage{url}            % simple URL typesetting

\usepackage{booktabs}       % professional-quality tables
\usepackage{amsfonts}       % blackboard math symbols
\usepackage{nicefrac}       % compact symbols for 1/2, etc.
\usepackage{microtype}      % microtypography

\usepackage[left=1in,right=1in,top=1in,bottom=1in]{geometry}

\usepackage{dsfont}
\usepackage{amsmath,amssymb,amsthm}
\usepackage{color}
\usepackage{graphicx}
\usepackage{array}

\newtheorem{proposition}{Proposition}

\newtheorem{remark}{Remark}

\usepackage{algorithm, algorithmic, multirow}
\newcommand{\X}{\mathcal{X}}
\newcommand{\W}{\mathcal{W}}
\newcommand{\R}{\mathbb{R}}
\newcommand{\E}{\mathbb{E}}
\newcommand{\mO}{\mathcal{O}}

\newcommand{\acqf}{\rho\textnormal{KG}}
\newcommand{\acqfapx}{\rho\textnormal{KG}^{apx}}

\title{Bayesian Optimization of Risk Measures \footnote{To appear in NeurIPS 2020.}}

\author{Sait Cakmak\textsuperscript{1}, Raul Astudillo\textsuperscript{2}, Peter Frazier\textsuperscript{2}, and Enlu Zhou\textsuperscript{1} \\
   \textsuperscript{1}School of Industrial and Systems Engineering,
   Georgia Institute of Technology \\
   \textsuperscript{2}School of Operations Research and Information Engineering,
   Cornell University 
}

\begin{document}

\maketitle

\begin{abstract}
  We consider Bayesian optimization of objective functions of the form $\rho[ F(x, W) ]$, where $F$ is a black-box expensive-to-evaluate function and $\rho$ denotes either the VaR or CVaR risk measure, computed with respect to the randomness induced by the environmental random variable $W$. Such problems arise in decision making under uncertainty, such as in portfolio optimization and robust systems design. We propose a family of novel Bayesian optimization algorithms that exploit the structure of the objective function to substantially improve sampling efficiency. Instead of modeling the objective function directly as is typical in Bayesian optimization, these algorithms model $F$ as a Gaussian process, and use the implied posterior on the objective function to decide which points to evaluate. We demonstrate the effectiveness of our approach in a variety of numerical experiments.
\end{abstract}

\section{Introduction}
Traditional Bayesian optimization (BO) has focused on problems of the form $\min_x F(x)$, or more generally,
$\min_x \E\left[F(x,W)\right]$, where $F$ is a time-consuming black box function that does not provide derivatives, and $W$ is a random variable.
This has seen an enormous impact, and has expanded from hyper-parameter tuning \cite{Snoek2012ML, zller2019benchmark} to more sophisticated applications such as drug discovery and robot locomotion \cite{Negoescu2011Drug, Jing2012Cardio, letham2018fbBO, rai2019using}. However, in many truly high-stakes settings, optimizing average performance is inappropriate: we must be risk-averse. In such settings, risk measures have become a crucial tool for quantifying risk. For instance, by law, banks are regulated using the Value-at-Risk \cite{Lopez1997Fed}.
Risk measures have also been used in cancer treatment planning \cite{an2017robust,lim2020risk}, healthcare operations \cite{Austin2003}, natural resource management \cite{Sethi2010Resource}, disaster management \cite{Merakli2019Disaster}, data-driven stochastic optimization \cite{Wu2018BRO}, and risk quantification in stochastic simulation \cite{Zhu2020input-uncertainty}.

In this work, we consider risk averse optimization of the form $\min_x \rho\left[F(x,W)\right]$, where $\rho$ is a {\it risk measure} that maps the probability distribution of $F(x,W)$ (induced by the randomness on $W$) onto a real number describing its level of risk.
We focus on the setting, where,
during the evaluation stage, $F$ can be evaluated for any $(x, w) \in \X \times \W$, e.g., using a simulation oracle. After the evaluation stage, we choose a decision $x^*$ to be implemented in the real world, nature chooses the random $W$, and the objective $F(x^*, W)$ is then realized.

When $F$ is inexpensive and has convenient analytic structure, optimizing against a risk measure is well understood \cite{Ruszczynski2006Risk, BenTal2009RO, Bertsimas2011DRO, cakmak2020solving}.
However, when $F$ is {\it expensive-to-evaluate}, {\it derivative-free}, or is a {\it black box}, the existing literature is inadequate in answering the problem.
A naive approach would be to use the standard BO algorithms with observations of $\rho[F(x,W)]$. However, a single evaluation of the objective function, $\rho[F(\cdot,W)]$, requires multiple evaluations of $F$, which can be prohibitively expensive when the evaluations of $F$ are expensive. For example, if each evaluation of $F$ takes one hour, and we need $10^2$ samples to obtain a high-accuracy estimate of $\rho[F(x,W)]$, a single evaluation of the objective function would require more than four days.
Moreover, if we do not obtain a high-accuracy estimate, estimators are typically biased \cite{TRINDADE2007risk} in a way that is unaccounted for by traditional Gaussian process regression. 
For the risk measure Value-at-Risk, the recent work of \cite{torossian2020bayesian} addresses this issue by modeling $\text{VaR}_{\alpha}[F(x, W)]$ using individual observations of $F(x, w)$. However, their method only chooses the $x$ to evaluate, and $w$ is randomly set according to its environmental distribution.
As evidenced by \cite{janusevskis2013mean, BQO}, jointly selecting $x$ and $w$ to evaluate offers significant improvements to query efficiency.
This ability, unlocked when evaluations are made using a simulation oracle, is leveraged by the methods we introduce here.

As an example of the benefits of choosing $w$ intelligently,
consider a function $F$ that is monotone in $w$. To optimize VaR of $F$, we only need to evaluate a single value of $w$, the one that corresponds to the VaR, and evaluating any other $w$ is simply inefficient. 
In a black-box setting, we would not typically know that $F$ was monotone but, by modeling $F$ directly, we can discover the regions of $w$ that matter and focus most of our effort into selectively evaluating $w$ from these regions.

In this paper, 
we focus on two risk measures commonly used in practice, Value-at-Risk (VaR) and Conditional Value-at-Risk (CVaR);
and develop a novel approach that overcomes the aforementioned challenges. Our contributions are summarized as follows:
\begin{itemize}

    \item
    To the best of our knowledge, our work is the first to consider BO of risk measures 
    while leveraging the ability to choose $x$ {\it and} $w$ at query time.
    The selection of $w$ enables efficient search of the solution space, and is a significant contributor to the success of our algorithms.
    \item We provide a novel one-step Bayes optimal algorithm $\acqf$, and a fast approximation $\acqfapx$ that performs well in numerical experiments; significantly improving the sampling efficiency over the state-of-the-art BO methods.
    
    \item We combine ideas from different strands of literature to derive 
    gradient estimators for efficient optimization of $\acqf$ and $\acqfapx$, which are shown to be asymptotically unbiased and consistent. 
    
    \item To further improve the computational efficiency, we propose a two time scale optimization approach, that is broadly applicable for optimizing acquisition functions whose computation involves solving an inner optimization problem.
\end{itemize}

The remainder of this paper is organized as follows. Section \ref{section-background} provides a brief background on BO and risk measures. Section \ref{section-setup} formally introduces the problem setting. Section \ref{section-statistical-model} introduces the statistical model on $F$, a Gaussian process (GP) prior, and explains how to estimate VaR and CVaR of a GP. Section \ref{section-acqf} introduces $\acqf$, a knowledge gradient type of acquisition function for optimization of VaR and CVaR, along with a cheaper approximation, $\acqfapx$, and efficient optimization of both acquisition functions. Section \ref{section-numerical} presents numerical experiments demonstrating the performance of the algorithms developed here. Finally, the paper is concluded in Section \ref{section-conclusion}.

\section{Background}\label{section-background}

\subsection{Bayesian optimization}
BO is a framework for global optimization of expensive-to-evaluate, black-box objective functions \cite{Frazier2018Tutorial}, whose origins date back to the seminal work of \cite{kushner1964new}. Among the various types of acquisition functions in BO \cite{Jones1998EI,Scott2011KG,hernandez2014predictive}, our proposed acquisition functions can be catalogued as knowledge gradient methods, which have been key to extending BO beyond the classical setting, allowing for parallel evaluations \cite{Wu2016ParallelKG}, multi-fidelity observations \cite{poloczek2017multi}, and gradient observations \cite{wu2017gradient}.

Within the BO literature, a closely related work to ours is \cite{BQO}, 
which studies the optimization of $\E_W[F(x, W)]$, while jointly selecting $x$ and $w$ to evaluate.
Due to linearity of the expectation, the GP prior on $F(x, w)$ translates into a GP prior on $\E_W[F(x, W)]$, a fact that is leveraged to efficiently compute the one-step Bayes optimal policy, also known as the knowledge gradient (KG) policy. Although it is not the focus of this study, since CVaR at risk level $\alpha = 0$ is the expectation, our methods can be directly applied for solving the expectation problem, and the resulting algorithm is equivalent to the algorithm proposed in \cite{BQO}. 

Our work also falls within a  strand of the BO literature that aims to find solutions that are risk-averse to the effect of an unknown environmental variable \cite{marzat2013, Bogunovic2018Adversarial, torossian2020bayesian, Kirschner2020RobustBO, nguyen2020distributionally}. Worst-case optimization, also known as minimax or robust optimization, is considered by \cite{marzat2013} and \cite{Bogunovic2018Adversarial}, whereas \cite{Kirschner2020RobustBO} and \cite{nguyen2020distributionally} consider distributionally robust optimization \cite{ rahimian2019distributionally}.
Within this line of research, \cite{torossian2020bayesian}, which studies the optimization of $\textnormal{VaR}_{\alpha}[F(x, W)]$, is arguably the most closely related to work. 
Like our statistical model, their model is also built using individual observations of $F(x,w)$ and not directly using observations of $\textnormal{VaR}_{\alpha}[F(x, W)]$. However, unlike our approach, their approach is only able to choose at which $x$ to evaluate. Our approach jointly chooses $x$ and $w$ to evaluate, which is critical when $\W$ is large. Moreover, our model allows for noisy evaluations of $F(x, w)$, which could introduce additional bias in their model.

\subsection{Risk measures}

We recall that a risk measure (cf. \cite{Artzner1999Coherent, ROCKAFELLAR2002CVaR, Delbaen2002, Rockafellar2013Risk}) is a functional that maps probability distributions onto a real number.
Risk measures offer a middle ground between the risk-neutral expectation operator and the worst-case performance measure, which is often more interpretable than expected utility \cite{Wu2018BRO}.
For a generic random variable $Z$, we often use the notation $\rho(Z)$ to indicate the risk measure evaluated on the {\it distribution} of $Z$.

The most widely used risk measure is VaR \cite{Jorion2007VaR},
which measures the maximum possible loss after excluding worst outcomes with a total probability of $1 - \alpha$, and is defined as
    $\text{VaR}_{\alpha}[Z] = \inf \{t: P_Z(Z \leq t) \geq \alpha \}$
where $P_Z$ indicates the distribution of $Z$.
Another widely used risk measure is CVaR,
which is the expectation of the worst losses with a total probability of $1 - \alpha$, and is given by
    $\text{CVaR}_{\alpha}[Z] = \mathbb{E}_Z[Z \mid Z \geq \text{VaR}_{\alpha}(Z)]$.

VaR and CVaR have been applied in a wide range of settings, including simulation optimization under input uncertainty \cite{Wu2018BRO, cakmak2020solving}, insurance \cite{Zympler2011Insurance} and risk management \cite{Jorion2010}. They are widely used in finance, and VaR is encoded in the Basel II accord \cite{Basel2012}. We refer the reader to \cite{Hong2014VaR-CVaR-Rev} for a broader discussion on VaR and CVaR.

\section{Problem setup} \label{section-setup}
We consider the optimization problem
\begin{equation}\label{generic-problem}
    \min_{x \in \X} \rho\left[F(x, W) \right],
\end{equation}
where $\X \subset \R^{d_\X}$ is a simple compact set, e.g., a hyper-rectangle; $W$ is a random variable with probability distribution $\mathbb{P}_W$ and compact support $\W \subset \R^{d_\W}$; and $\rho$ is a known risk measure, mapping the random variable $F(x, W)$ (induced by $W$) to a real number. We assume that $F: \X \times \W \rightarrow \R$ is a continuous black-box function whose evaluations are expensive, i.e., each evaluation takes hours or days, has significant monetary cost, or the number of evaluations is limited for some other reason. We also assume that evaluations of $F$ are either noise-free or observed with independent normally distributed noise with known variance.

We emphasize that the risk measure $\rho$ is only over the randomness in $W$, and $\rho\left[F(x,W)\right]$ is calculated holding $F$ fixed. For example, if $\rho$ is VaR, and $W$ is a continuous random variable with density $p(w)$, then this is explicitly written as:
\begin{equation}
    \text{VaR}_{\alpha}[F(x, W)] = \inf \left\{t: \int_{\W} \mathds{1}\{F(x,w) \leq t\} p(w)\,dw \geq \alpha \right\}
\end{equation}

Later, we will model $F$ as being drawn from a GP, but we will continue to compute $\rho\left[F(x,W)\right]$ only over the randomness in $W$. Thus, this quantity is a function of $F$ and $x$, and will be random with a distribution induced by the distribution on $F$ (and more precisely by the distribution on $F(x,\cdot)$).

\section{Statistical model} \label{section-statistical-model}
We assume a level of familiarity with GPs and refer the reader to \cite{Rasmussen2006GP} for details.
We place a GP prior on $F$, specified by a mean function $\mu_0: \X \times \W \rightarrow\R$ and a positive definite covariance function $\Sigma_0:(\X \times \W)^2\rightarrow\R^+$, and assume that queries of $F$ are of the form $y_i = F(x_i, w_i) + \epsilon_i$, where $\epsilon_i\sim N(0,\sigma^2)$ is independent across evaluations. The posterior distribution on $F$ given the history after $n$ evaluations, $\mathcal{F}_n := \{(x_i, w_i), y_i\}_{i=1}^n$, is again a GP with mean and covariance functions $\mu_n$ and $ \Sigma_n$, which can be computed in closed form in terms of $\mu_0$ and $\Sigma_0$.

The GP posterior distribution on $F$ implies a posterior distribution on the mapping $x \mapsto \rho[F(x, W)]$. In contrast with the case where $\rho$ is the expectation operator, this distribution is, in general, non-Gaussian, rendering computations more challenging. As we discuss next, however, quantities of  interest can still be computed following a simple Monte Carlo (MC) approach. Specifically, we discuss how to compute the posterior mean of $\rho[F(x, W)]$, i.e., $\E_n[\rho[F(x, W)]]$, where $\E_n$ denotes the conditional expectation  given $\mathcal{F}_n$.

Our approach to estimating the value of $\E_n[\rho[F(x, W)]]$ builds on samples of $[F(x, w): w \in \W]$ drawn from the posterior, and the corresponding implied values of $\rho[F(x, W)]$. For the purposes of this discussion, we assume  $\W$ is finite and small, say  $\W = \{w_1,\ldots, w_L\}$, and that $\mathbb{P}_W$ is uniform over $\W$. When the cardinality of $\W$ is finite but large or $\mathbb{P}_W$ is continuous, we instead use a set of $L$ i.i.d. samples drawn from $\mathbb{P}_W$.

Denote $w_{1:L} = [w_1, \ldots, w_L]$ and  $F(x, w_{1:L}) = [F(x,w_1), \ldots, F(x,w_L)]$. The time-$n$ joint posterior distribution of $F(x, w_{1:L})$ is normal with mean vector $\mu_n(x, w_{1:L})$ and covariance matrix $\Sigma_n(x, w_{1:L}, x, w_{1:L})$.  
A sample from this distribution can be obtained via the reparameterization trick \cite{wilson2018maximizing}, i.e., as $\mu_n(x, w_{1:L}) + C_n(x, w_{1:L})Z$, where is $C_n(x, w_{1:L})$ is the Cholesky factor of $\Sigma_n(x, w_{1:L}, x, w_{1:L}))$ and $Z$ is drawn from the $L$-variate standard normal distribution.

Let $\widehat{F}(x, w_{1:L})$ denote a realization of $F(x, w_{1:L})$, computed as described above. The corresponding MC estimate of $\rho[F(x, W)]$ can be calculated as the empirical risk measure corresponding to this realization, which, under the assumption that $\mathbb{P}_\W$ is uniform over $\W$, can obtained by ordering the coordinates of $\widehat{F}(x,w_{1:L})$ so that $\widehat{F}(x, w_{(1)}) \leq \widehat{F}(x, w_{(2)}) \leq \ldots \leq \widehat{F}(x, w_{(L)})$, and letting
\begin{equation} \label{eq-rho-samples}
    \widehat{v}(x) := \widehat{F}(x, w_{(\lceil L \alpha \rceil)}) \quad \text{ and } \quad  \widehat{c}(x) := \frac{1}{\lceil L (1 - \alpha) \rceil} \sum_{j = \lceil L \alpha \rceil}^L \widehat{F}(x, w_{(j)}),
\end{equation}
be the empirical VaR and CVaR respectively (cf. \cite{Hong2014VaR-CVaR-Rev}), where $\lceil \cdot \rceil$ is the ceiling operator. 
This can be easily extended to non-uniform discrete 
$\mathbb{P}_W$ by accounting for the probability mass of each $w \in \W$.

Finally, if  $\widehat{v}^j(x)$ and $\widehat{c}^j(x)$, $j=1,\ldots, M$, are samples of $\text{VaR}_{\alpha}[F(x, W)]$ and $\text{CVaR}_{\alpha}[F(x, W)]$, obtained as described above. Then, MC estimates of $\E_n[\text{VaR}_{\alpha}[F(x, W)]]$ and $\E_n[\text{CVaR}_{\alpha}[F(x, W)]]$ are given by $\frac{1}{M} \sum_{j=1}^M \widehat{v}^j(x)$ and $\frac{1}{M} \sum_{j=1}^M \widehat{c}^j(x)$, respectively.

\section{The $\acqf$ acquisition function} \label{section-acqf}

As is standard in the BO literature, our algorithm's search is guided by an acquisition function, whose maximization indicates the next point to evaluate. Our proposed acquisition function  generalizes the well-known knowledge gradient acquisition function \cite{Scott2011KG}, which has been generalized to other settings such as parallel and multi-fidelity optimization \cite{Wu2016ParallelKG,poloczek2017multi}.

Before formally introducing our acquisition function, we note that, in our setting, choosing a decision $x$ to be implemented after the evaluation stage is complete, is not a straightforward task. If the cardinality of $\W$ is large, then the chances are $\rho[F(x, W)]$ will not be known exactly as this would require evaluating $F(x, w)$ for all $w\in\W$. A common approach in such scenarios is to choose the decision with best expected objective value according to the posterior distribution (c.f. \cite{Frazier2018Tutorial}), 
% i.e., the solution of
% \begin{equation}
% \label{eqn:objval}
%     \min_{x\in\X} \mathbb{E}_N[ \rho[F(x, W)]].
% \end{equation}
\begin{equation}
\label{eqn:objval}
    \min_{x\in\X} \mathbb{E}_N[ \rho[F(x, W)]].
\end{equation}
Having defined the choice of the decision $x$ to be implemented after the evaluation stage is complete, we can now introduce our acquisition function, the knowledge gradient for risk measures. We motivate this acquisition function by noting that, if we had to choose a decision with the information available at time $n$, the expected objective value we would get according to equation (\ref{eqn:objval}) is simply  $\rho_n^* := \min_{x\in\X} \mathbb{E}_n[ \rho[F(x, W)]].$ On the other hand, if we were allowed to make one additional evaluation, the expected objective value we would get is 
% $\min_{x\in\X} \mathbb{E}_{n+1}[ \rho[F(x, W)]]$.
$\rho_{n+1}^*:= \min_{x\in\X} \mathbb{E}_{n+1}[ \rho[F(x, W)]]$.
Therefore,
% \begin{equation*}
%     \min_{x\in\X} \mathbb{E}_n[ \rho[F(x, W)]] - \min_{x\in\X} \mathbb{E}_{n+1}[ \rho[F(x, W)]]
% \end{equation*}
$\rho_n^* - \rho_{n+1}^*$
measures improvement due to making one additional evaluation.

We emphasize that, given the information available at time $n$,
$\rho_{n+1}^*$
% $\min_{x\in\X} \mathbb{E}_{n+1}[ \rho[F(x, W)]]$ 
is random due to its dependence on the yet unobserved $(n+1)$-st evaluation. The knowledge gradient for risk measures acquisition function is defined as the expected value of $\rho_n^* - \rho_{n+1}^*$ given the information available at time $n$ and the 
next point $(x, w)$ to evaluate, termed the "candidate" from here on:
\begin{equation}\label{rhokg-acq}
\begin{aligned}
    \acqf_n(x, w) ={} &\mathbb{E}_n \left[ \rho_n^* - \rho_{n+1}^* \mid (x_{n+1}, w_{n+1}) = (x, w)\right]
\end{aligned}
\end{equation}
Our algorithm sequentially chooses the next point to evaluate as the candidate that maximizes (\ref{rhokg-acq}), and is, by construction, one-step Bayes optimal.

\subsection{Optimization of $\acqf$}

In this section, we discuss how to evaluate and optimize $\acqf$ using a sample average approximation (SAA) approach \cite{Kim2015SAA}. In a nutshell, this approach works by constructing a MC approximation of $\acqf_n(x,w)$, that is deterministic given a finite set of \textit{base samples} not depending on the candidate $(x,w)$. Such an approximation can be optimized using deterministic optimization methods. This is usually faster than optimizing the original acquisition function with stochastic optimization techniques. Below, we discuss how to construct this SAA. Moreover, we show that its gradients can be readily computed, thus allowing the use of higher-order optimization methods. A more detailed discussion of our approach to optimize $\acqf$ can be found in the supplement.  

We begin by noting that $\rho_n^*$ does not depend on the candidate being evaluated. Therefore, maximizing $\acqf$ is equivalent to solving
\begin{equation}
\label{eq-acqf-equiv}
    \max_{(x,w)\in\X\times\W}\mathbb{E}_n \left[ -\rho_{n+1}^{*} \mid (x_{n+1}, w_{n+1}) = (x, w)\right].
\end{equation}
The first step in building an SAA of (\ref{eq-acqf-equiv}) is to draw $K$ \textit{fantasy} samples from the time-$n$ posterior distribution on $y_{n+1}$, which, conditioned on $(x_{n+1}, w_{n+1}) = (x, w)$, is Gaussian with mean $\mu_n(x, w)$ and variance $\Sigma_n(x, w, x, w)$. 
Using the reparameterization trick, these samples can be obtained as $\mu_n(x, w) + \Sigma_n(x, w, x, w)Z^i$, $i=1,\ldots, K$, where the \textit{base samples} $Z^1,\ldots, Z^K$ are drawn 
% \raedit{independently(?)} - qMC is not independent. Let's not restrict. 
from a standard normal distribution. 
These samples give rise to $K$ \textit{fantasy} GP models of the posterior distribution at time $n+1$, obtained by conditioning the GP model on the event $(x_{n+1}, w_{n+1}) = (x, w)$ and $y_{n+1} = \mu_n(x, w) + \Sigma_n(x, w, x, w)Z^i$, i.e., by adding $\{(x, w), \mu_n(x, w) + \Sigma_n(x, w, x, w)Z^i\}$ as the hypothetical $(n+1)$-st observation. 

For each fantasy GP model $i$, an MC estimate of $\E_{n+1}[\rho[F(x^i, W)]]$, $x^i\in\X$, can be constructed by averaging samples as described in Section \ref{section-statistical-model}. Let $\mathfrak{r}^{ij}(x^i)$ denote the $j$-th such sample corresponding to the $i$-th fantasy GP model, where the dependence of $\mathfrak{r}^{ij}(x^i)$ on the candidate $(x,w)$ and $Z^i$ is made implicit. 
Additional base samples 
needed to define $\mathfrak{r}^{ij}(x^i)$ are generated once and held fixed so that it becomes a deterministic function of $x^i,(x,w)$.
The SAA of (\ref{eq-acqf-equiv}) is then given by
\begin{equation} \label{eq-acqf-mc-estimate}
    \max_{(x,w)\in\X\times\W} - \frac{1}{K} \sum_{i=1}^K \min_{x^{i}\in\X}\frac{1}{M} \sum_{j=1}^M \mathfrak{r}^{ij}(x^{i}).
\end{equation}

In the supplement, we show that the gradient of $\mathfrak{r}^{ij}$ with respect to $x^i$, denoted $\nabla_{x^i} \mathfrak{r}^{ij}(x^i)$, can be 
computed explicitly 
as $\nabla_{x^i} \widehat{F}^{ij}(x^i, w_{(\lceil L \alpha \rceil)})$ and $\frac{1}{\lceil L (1 - \alpha) \rceil} \sum_{j = \lceil L \alpha \rceil}^L \nabla_{x^i} \widehat{F}^{ij}(x^i, w_{(j)})$ for VaR and CVaR respectively,
where this notation is defined explicitly in the supplement.
We use these gradients within the LBFGS algorithm \cite{Zhu1997LBFGS} to solve the inner optimization problems in (\ref{eq-acqf-mc-estimate}). 
Moreover, we show that, under mild regularity conditions, the envelope theorem (\cite{milgrom2002envelope}, Corollary 4) can be used to express the gradient of the objective in (\ref{eq-acqf-mc-estimate}) as
\begin{equation}
\label{eq-outer-gradient-estimators}
   - \frac{1}{K} \sum_{i=1}^K \frac{1}{M} \sum_{j=1}^M \nabla_{(x, w)}\mathfrak{r}^{ij}(x_*^i), 
\end{equation}
where $\nabla_{(x, w)}\mathfrak{r}^{ij}(x_*^i)$ denotes the gradient of $\mathfrak{r}^{ij}$ with respect to $(x,w)$ evaluated at $x_*^i$, and $x_*^i$ is a solution to the $i$-th  inner optimization problem in (\ref{eq-acqf-mc-estimate}). Again, we use these gradients within LBFGS to solve (\ref{eq-acqf-mc-estimate}). In addition, we also show that the above gradients are asymptotically unbiased and consistent gradient estimators as $K, L, M \rightarrow \infty$. Therefore, $\acqf$ can be maximized using (multi-start) stochastic gradient ascent (SGA), following an approach similar to a proposal in \cite{wu2017gradient}. 

\begin{proposition}
    Under suitable regularity conditions,  (\ref{eq-outer-gradient-estimators}) is an asymptotically unbiased and consistent estimator of the gradient of $\acqf$
    as $K, L, M \rightarrow \infty$.
\end{proposition}

A formal statement of the proposition and its proof is given in the supplement for the case of $d^\W = 1$.
It is also shown that selecting the $n$-th candidate to evaluate using the $\acqf$ algorithm, including the training of the GP model, has a computational complexity of $\mO(Q_1 n^3 + Q_2  Q_3 K L [n^2 + L n + L^2 + ML])$, where $Q_1, Q_2, Q_3$ are the number of LBFGS \cite{Zhu1997LBFGS} iterations performed for training the GP model, and the outer and inner optimization loops respectively.

\begin{remark}
This and the preceding sections are explained using MC estimators.  The same approach works using quasi-MC estimators, obtained by generating $Z$ in reparameterization (see Section \ref{section-statistical-model}) using Sobol sequences \cite{OWEN1998Sobol}. In practice we use quasi-MC because it improves computationally over a simple MC approach. 
\end{remark}

\subsection{The $\acqfapx$ approximation} \label{section-kgcp}

The $\acqf$ algorithm is computationally intensive, as it requires solving a nested non-convex optimization problem. In a wide range of settings, the sampling efficiency it provides justifies its computational cost, but in certain not-so-expensive settings, a faster algorithm is desirable. 

Inspired by the EI \cite{Jones1998EI} and KGCP  \cite{Scott2011KG} acquisition functions, we propose $\acqfapx$, which replaces the inner optimization problem of $\acqf$ with a much simpler one. In $\acqfapx$, the inner optimization is restricted to the points $x$ that have been evaluated for at least one $w$, 
denoted by $\widetilde{\X}_n = \{x_{1:n}\}$,
and the resulting value of the optimization problem is $\widetilde{\rho}^*_n = \min_{x \in \widetilde{\X}_n} \mathbb{E}_n[ \rho[F(x, W)] ]$.
The intuition behind this is that the GP model is an extrapolation of the data. Thus, these points carry an immense amount of information on the GP model and the posterior objective, which makes them an ideal set of candidates to consider.
% In the following, the set $\widetilde{\X} = \{x_{1:n}\}$ is the $x$ components of all the solutions evaluated so far, in addition the set $\widetilde{\X}^+ = \{x_{1:n}, x_{n+1}\}$ includes the $x$ component of the current candidate in consideration.
% \begin{equation}\label{approx-acqf}
%     \acqfapx(x, w) =\mathbb{E}_n \left[ \min_{x\in\widetilde{\X}^n} \mathbb{E}_n[ \rho[F(x, W)] ] - \min_{x\in\widetilde{\X}^{n+1}} \mathbb{E}_{n+1} [\rho[F(x, W)]] \mid (x_{n+1}, w_{n+1}) = (x, w)\right].
% \end{equation}
The resulting approximation to $\acqf$ is,
\begin{equation}\label{approx-acqf}
    \acqfapx(x, w) =\mathbb{E}_n \left[ \widetilde{\rho}^*_n - \widetilde{\rho}^*_{n+1} \mid (x_{n+1}, w_{n+1}) = (x, w)\right].
\end{equation}

We note that, like $\acqf$, $\acqfapx$ has an appealing one-step Bayes optimal interpretation; the maximizer of $\acqfapx$ is the one-step optimal point to evaluate if we restrict the choice of the decision to be implemented, $x$, among those that have been evaluated for at least one environmental condition, $w$.

\subsection{Two time scale optimization} \label{section-tts}

In this paper, we introduce two acquisition functions, $\acqf$ and $\acqfapx$. These acquisition functions share a nested structure, making their maximization computationally challenging. Here, we describe a novel two time scale optimization approach for reducing this computational burden.

Since the posterior mean and kernel are continuous functions of the data, if we fix the base samples used to generate the fantasy model and the GP sample path, 
a small perturbation to the candidate solution $(x, w)$ results in only a slight shift to the sample path. 
Thus, the optimal solutions to the inner problems, obtained using the previous candidate solution, should remain within a small neighborhood of a current high quality local optimal solution (and likely of the global solution). 
% then continuity of the posterior mean and kernel causes 
% a small perturbation to the candidate solutions to shift the sample path only slightly. 
% Thus, optimal solutions obtained for the previous candidate should be in a small neighborhood of the current optimal solutions. 
We can thus use the inner solutions obtained in the previous iteration to obtain a good approximation of the acquisition function value and its gradient for the current candidate. 
We utilize this observation by solving the inner optimization problem once every $T\approx10$ iterations, and using this solution to evaluate $\acqf$ for the remaining $T - 1$ iterations.
We refer to this approach as {\it two time scale optimization} and present an algorithmic description with more detail in the supplement.

% Our novel {\it two-time scale optimization approach}, explained in more detail the supplement, leverages this fact to reduce the cost of evaluating the acquisition function from the cost of solving many non-convex optimization problems to the cost of a single evaluation of said problems. 

The two time scale optimization approach outlined here is not limited to $\acqf$, and can be applied to other acquisition functions that require nested optimization, such as \cite{Scott2011KG, Wu2016qKG, BQO}. 
In numerical testing, the two-time-scale optimization approach did not affect the performance of either of our algorithms while offering significant computational savings. 
%Therefore, we use two-time-scale optimization for all the numerical examples presented in this paper.

\subsection{A visual analysis of the acquisition functions} \label{section-acqf-plots}

\begin{figure}[ht]
    \centering
    \hspace{0.02\linewidth}
    \includegraphics[width=0.3\linewidth]{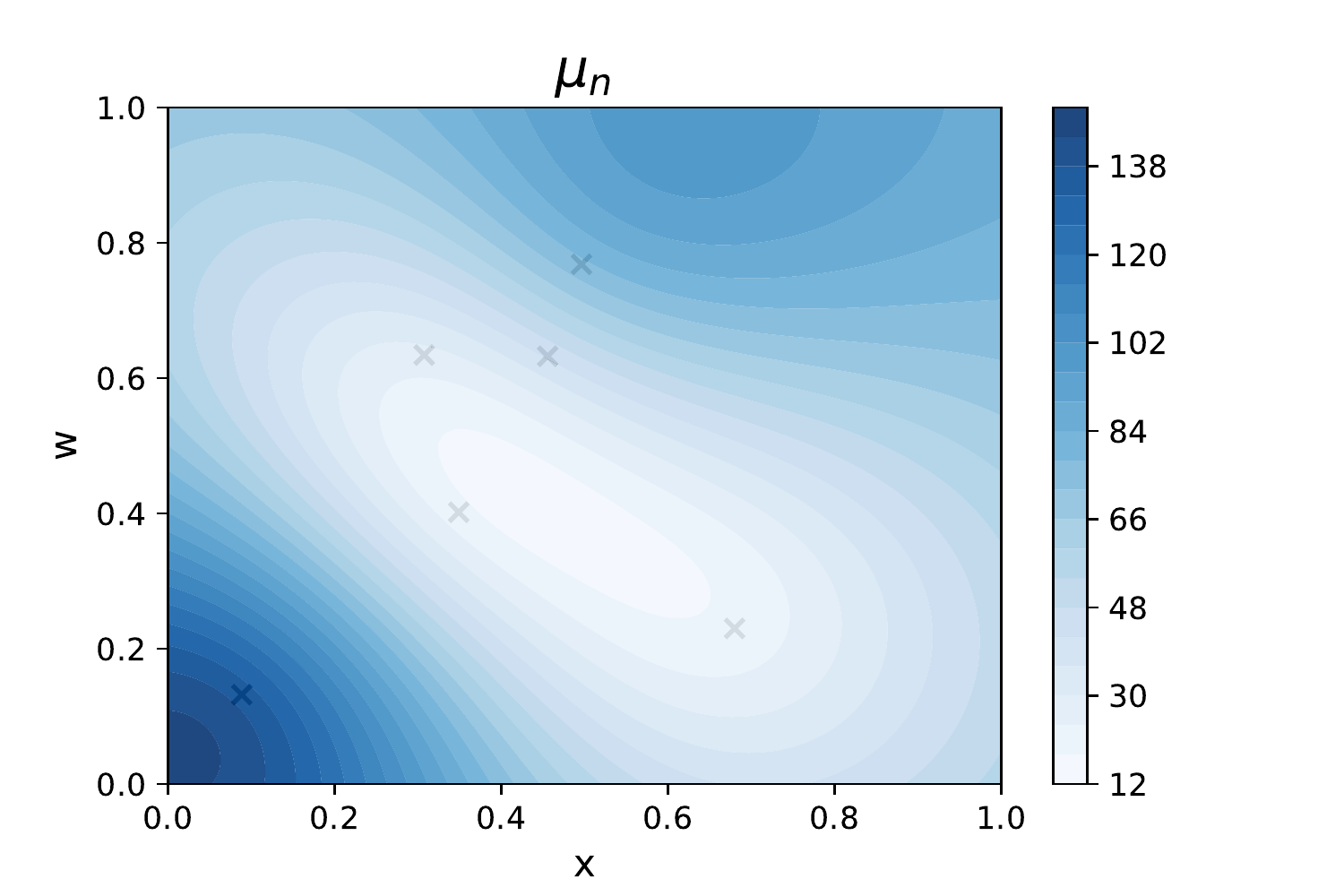}
    \includegraphics[width=0.3\linewidth]{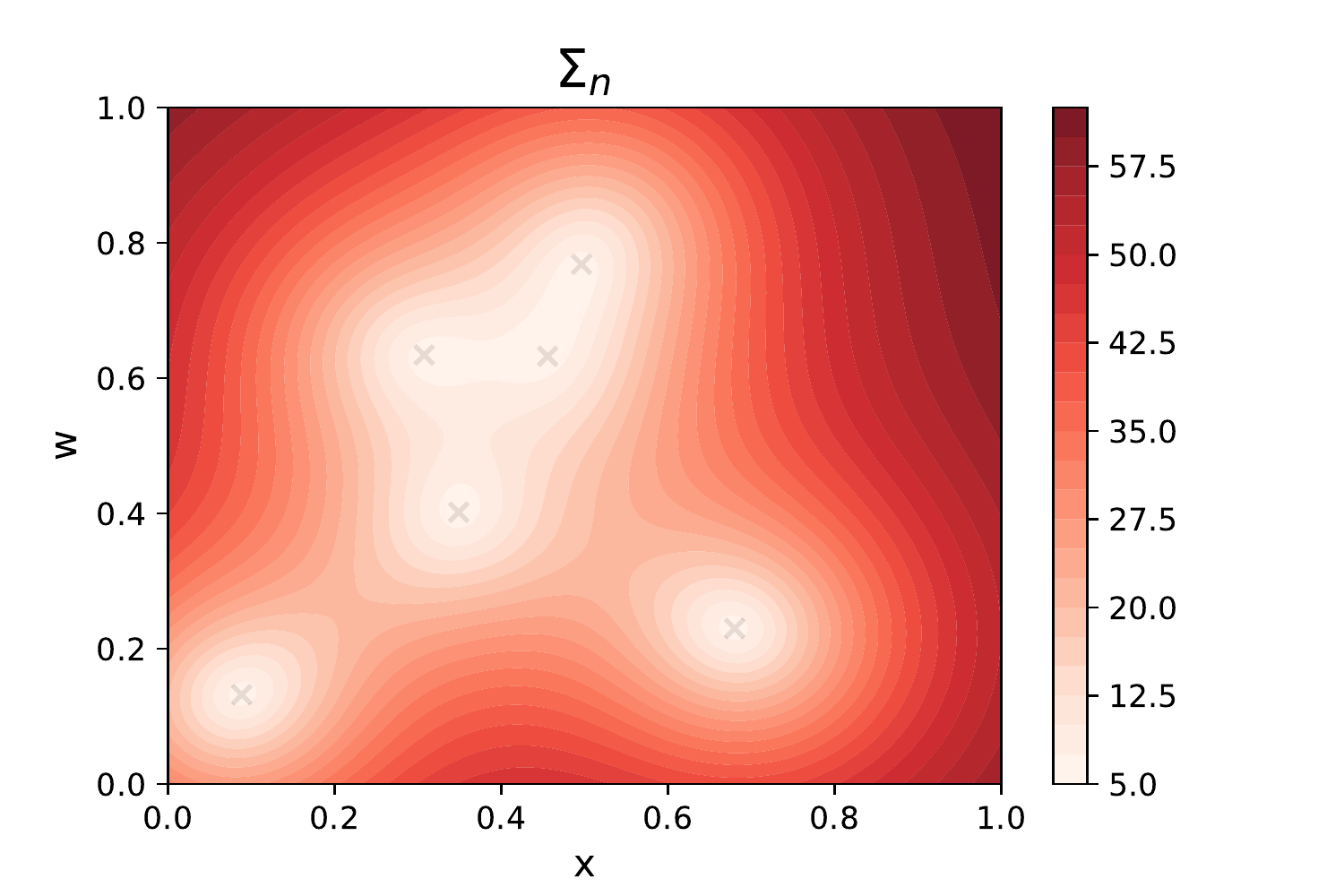}
    \includegraphics[width=0.3\linewidth]{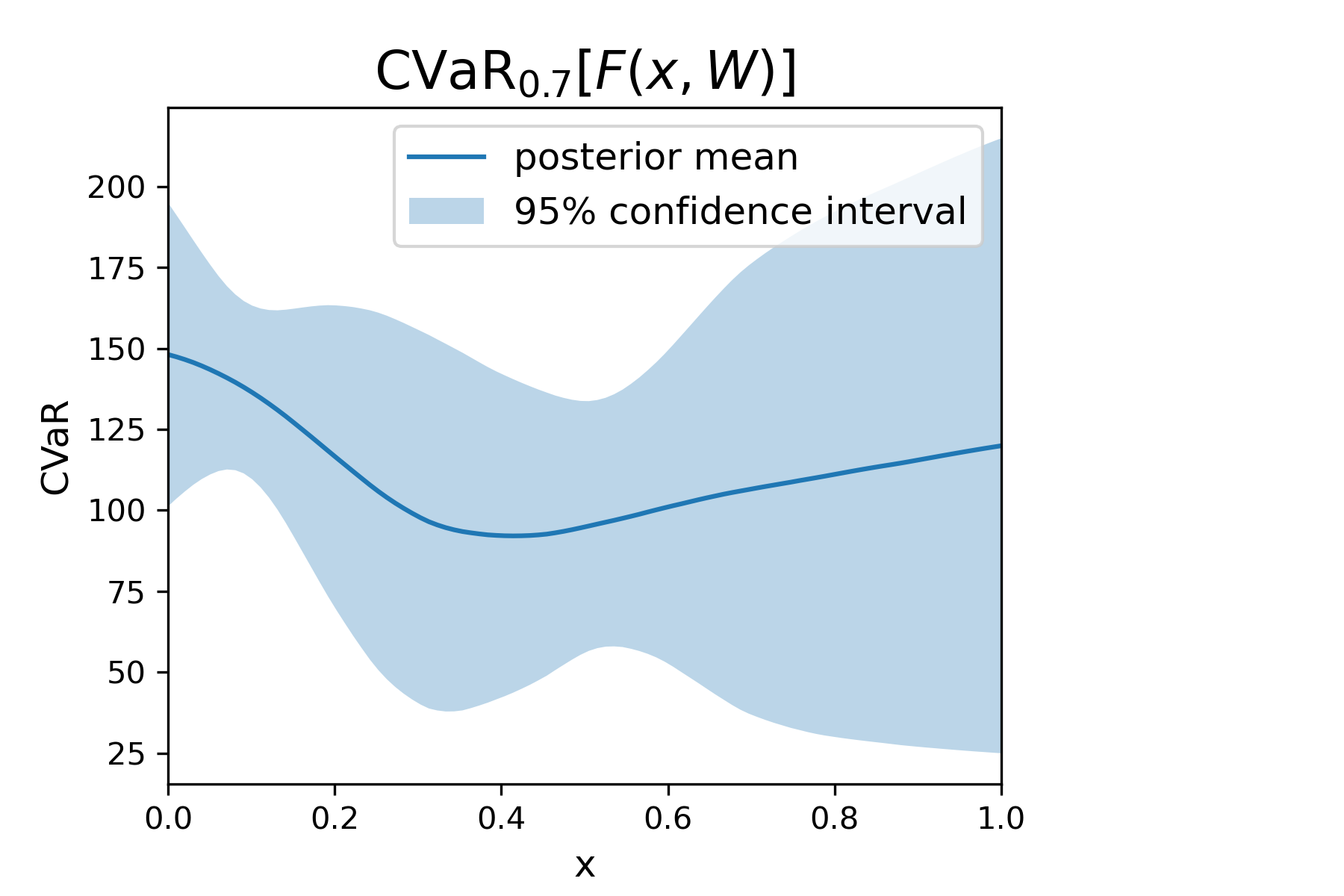} \\
    \includegraphics[width=0.3\linewidth]{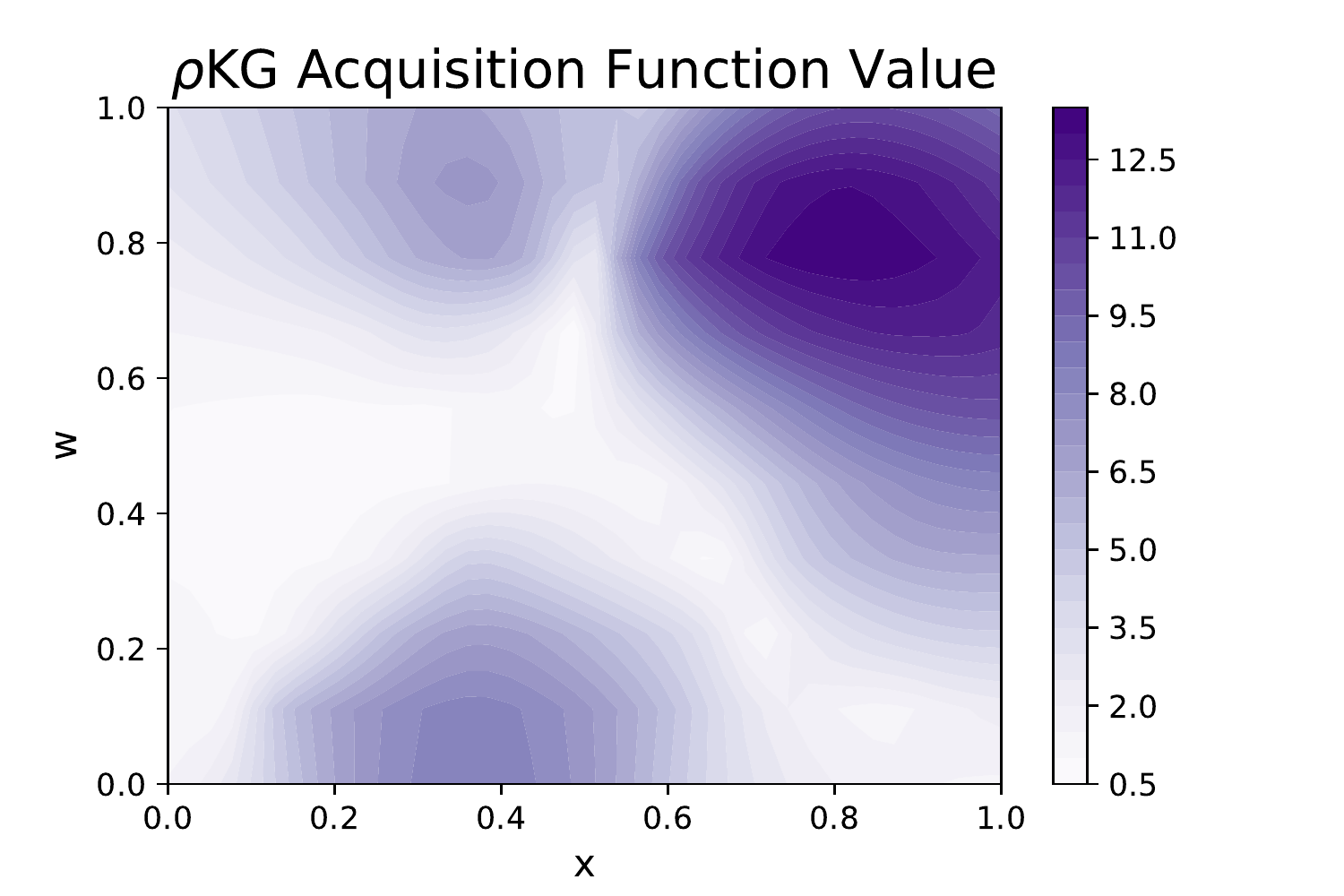}
    \includegraphics[width=0.3\linewidth]{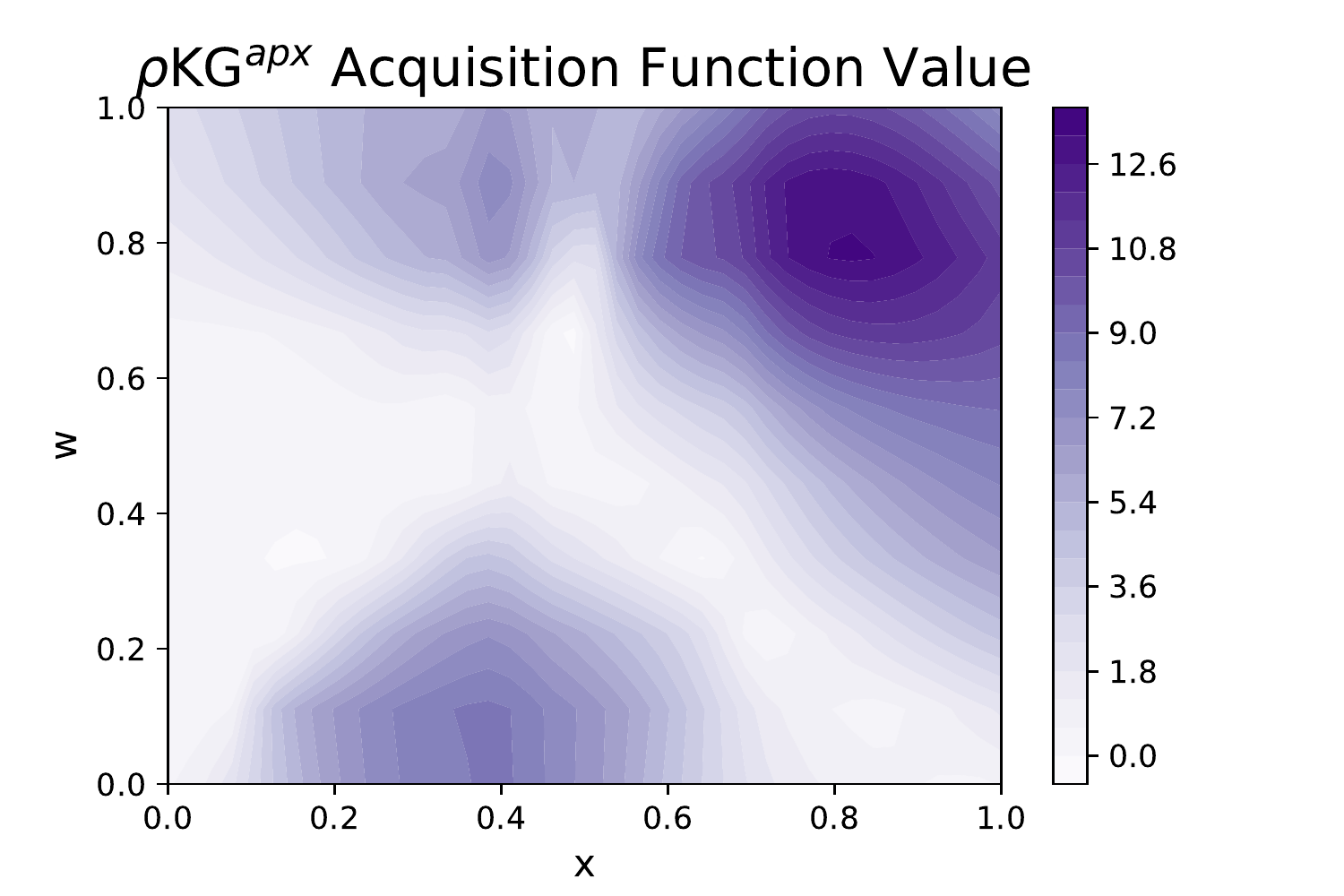}
    \caption{The top row shows the mean and variance functions of the posterior GP distribution on $F$, along with the implied (non-Gaussian) posterior distribution on  $\textnormal{CVaR}_{0.7}[F(\cdot,W)]$. The bottom row shows the $\acqf$ and $\acqfapx$ acquisition functions implied by the above statistical model.}
    \label{fig:gp-model-and-acqf-values}
\end{figure}

Figure \ref{fig:gp-model-and-acqf-values} plots a GP model based on 6 random samples taken from a 2-dim test function and the corresponding acquisition function values of $\acqf$ and $\acqfapx$ over $\X \times \W$. 
% We refer to the value of the acquisition function as the VOI, short for value of information. 
We use a CVaR objective with risk level $\alpha = 0.7$ and a uniform distribution over $\W = \{0/9, 1/9, \ldots, 9/9\}$. 

First, observe that the $\acqfapx$ plot closely resembles $\acqf$, supporting our claim that it is a good approximation. 
Second, the implied posterior on the objective shows a large uncertainty for larger values of $x$, and we expect a large $\acqf$ for these $x$ to encourage exploration. 
$\acqf$ plots show that this is indeed the case, while also showing a large variation along the $w$ axis. In this region of $x \in [0.8, 1.0]$, the posterior uncertainty is much larger for larger values of $w$. These $w$ also happen to have a posterior mean near likely $0.7$-quantiles of $F(x,\cdot)$, making them more informative about CVaR. Indeed, $\acqf$ prefers these $w$, reaching a maximum near $w=0.8$. 

Another area of interest is the promising region of $x \in [0.2, 0.4]$, as it contains the minimizer of the current posterior mean of the objective. We observe both a posterior mean substantially below the likely $0.7$-quantiles of $F(x,\cdot)$ and a low posterior uncertainty for $w \in [0.4, 0.6]$ in this region, making them less useful for estimating CVaR, and correspondingly low $\acqf$. However, the $w$ at the two ends of the range are more likely to be near an $0.7$-quantile and  have larger uncertainty. Thus, an observation from these $(x, w)$ would help pinpoint the exact location of the minimizer of the posterior expectation of the objective and thus are associated with larger $\acqf$.

We hope that these plots and accompanying discussion help the reader appreciate the value of considering the environmental variable, $w$, in designing the acquisition function.

\section{Numerical experiments}\label{section-numerical}

In this section, we present several numerical examples that demonstrate the sampling efficiency of our algorithms. We compare our algorithms with the Expected Improvement (EI, \cite{Jones1998EI}), Knowledge Gradient (KG, \cite{Wu2016qKG}), Upper Confidence Bound (UCB), and Max Value Entropy Search (MES, \cite{Wang17MES}) algorithms. We use the readily available implementations from the BoTorch package \cite{balandat2019botorch}, and the default parameter values given there. 
The benchmark algorithms cannot utilize observations of $F(x, w)$ while optimizing $\rho[F(x, W)]$. Therefore, 
for these algorithms, we fit a GP on observations of $\rho[F(x, W)]$, which are obtained by evaluating a given $x \in \X$ for all $w \in \W$ (or a subset $\widetilde{\W}$) and then calculating the value of the risk measure on these samples. As a result, the benchmark algorithms require $|\W|$ (or $|\widetilde{\W}|$) samples per iteration whereas $\acqf$ and $\acqfapx$ require only one.
We could not compare with \cite{torossian2020bayesian} since the code was not available at the time of the writing of this paper.

We optimize each acquisition function using the LBFGS \cite{Zhu1997LBFGS} algorithm with $10 \times (d^\X + d^\W)$ restart points. The restart points are selected from $500 \times (d^\X + d^\W)$ raw samples using a heuristic. For the inner optimization problem of $\acqf$, we use $5 \times d^\X$ random restarts with $25 \times d^\X$ raw samples. For both $\acqf$ and $\acqfapx$, we use the two time scale optimization where we solve the inner optimization problem once every $10$ optimization iterations. $\acqf$ and $\acqfapx$ are both estimated using $K = 10$ fantasy GP models, and $M = 40$ sample paths for each fantasy model.

We initialize each run of the benchmark algorithms with $2 d^\X + 2$ starting points from the $\X$ space, and the corresponding evaluations of $\rho[F(x,W)]$ obtained by evaluating $F(x, w)$ for each $w \in \W$ (or $\widetilde{\W}$, to be specified for each problem). The GP models for $\acqf$ and $\acqfapx$ are initialized using the equivalent number of $F(x, w)$ evaluations, with $(x, w)$ randomly drawn from $\X \times \W$. Further details on experiment settings is given in the supplement. The code for our implementation of the algorithms and experiments can be found at \url{https://github.com/saitcakmak/BoRisk}.

\subsection{Synthetic test problems}

The first two problems we consider are synthetic test functions from the BO literature. The first problem is the 4-dim Branin-Williams problem in \cite{Williams2000BW}. We consider minimization of both VaR and CVaR at risk level $\alpha=0.7$ with respect to the distribution of environmental variables $w = (x_2, x_3)$. The second problem we consider is the 7-dim $f_6(x_c, x_e)$ function from \cite{marzat2013}. We formulate this problem for minimization of CVaR at risk level $\alpha = 0.75$ with respect to the distribution of the $3$-dim environmental variable $x_e$. More details on these two problems can be found in the supplement.

\subsection{Portfolio optimization problem}
In this test problem, our goal is to tune the hyper-parameters of a trading strategy so as to maximize return under risk-aversion to random environmental conditions. We use CVXPortfolio \cite{cvxportfolio} to simulate and optimize the evolution of a portfolio over a period of four years using open-source market data. Each evaluation of this simulator returns the average daily return over this period of time under the given combination of hyper-parameters and environmental conditions. The  details  of this simulator can be found in Sections 7.1-7.3 of \cite{cvxportfolio}.

The hyper-parameters to be optimized are the risk and trade aversion parameters, and the holding cost multiplier over the ranges $[0.1,1000]$, $[5.5, 8.]$, and $[0.1,100]$, respectively. The environmental variables are the bid-ask spread and the borrow cost, which we assume are uniform over $[10^{-4},10^{-2}]$ and $[10^{-4},10^{-3}]$, respectively. For this problem, we use the VaR risk measure at risk level $\alpha=0.8$ We use a random subset $\widetilde{\W}$ of size $40$ for the inner computations of our algorithms, and a random subset of size $10$ is used for the evaluations of $\text{VaR}_{0.8}[F(x,W)]$ by the benchmark algorithms.

Since this simulator is indeed expensive-to-evaluate, with each evaluation taking around 3 minutes, evaluating the performance of the various algorithms becomes prohibitively expensive. Therefore, in our experiments we do not use the simulator directly. Instead, we build a surrogate function obtained as the mean function of a GP trained using evaluations of the actual simulator across $3,000$ points chosen according to a Sobol sampling design \cite{OWEN1998Sobol}.

\begin{figure}[ht]
    \centering
    \begin{tabular}{m{0.30\linewidth} m{0.30\linewidth} m{0.30\linewidth} }
      \includegraphics[width=\linewidth]{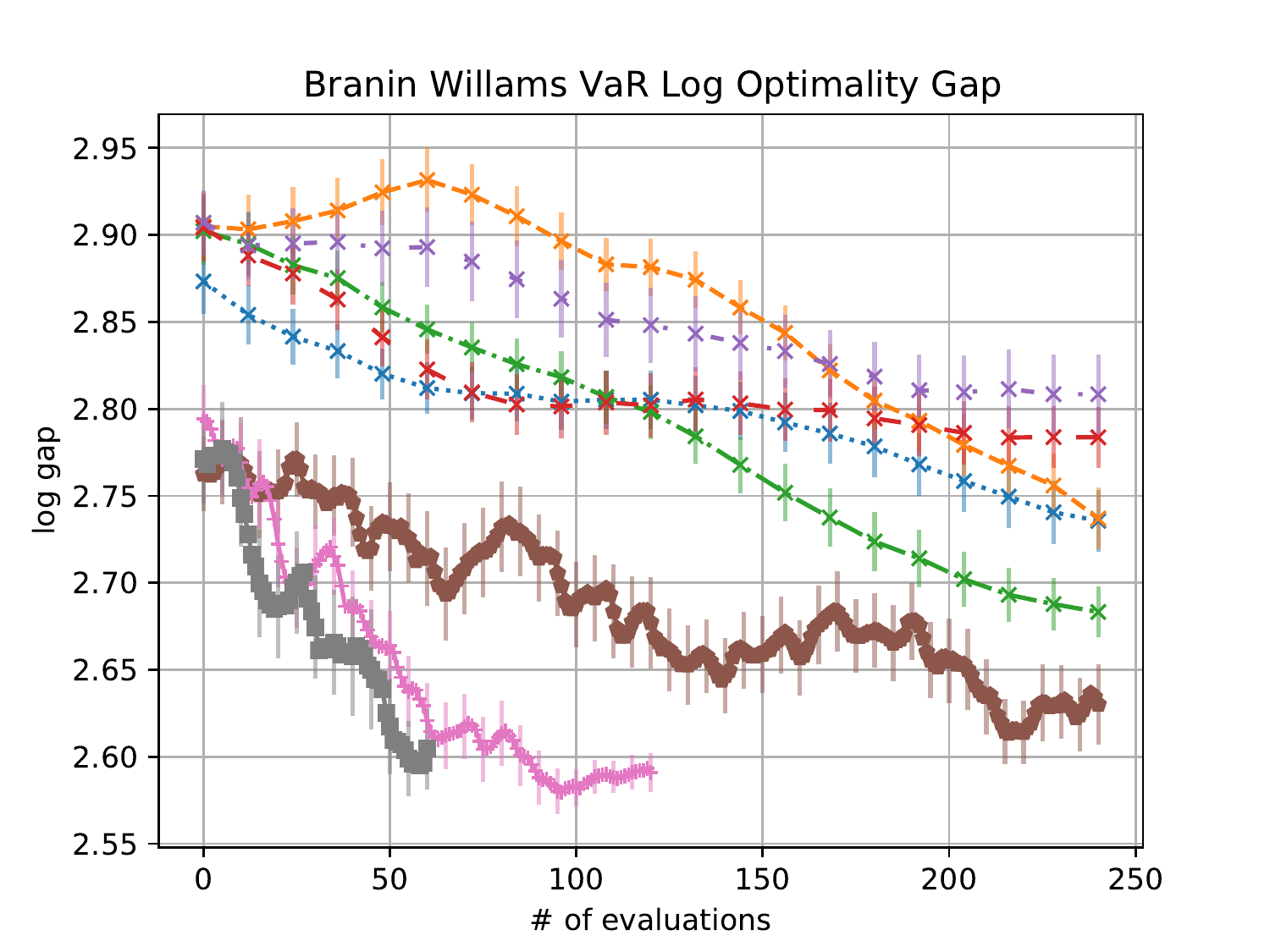}  & \includegraphics[width=\linewidth]{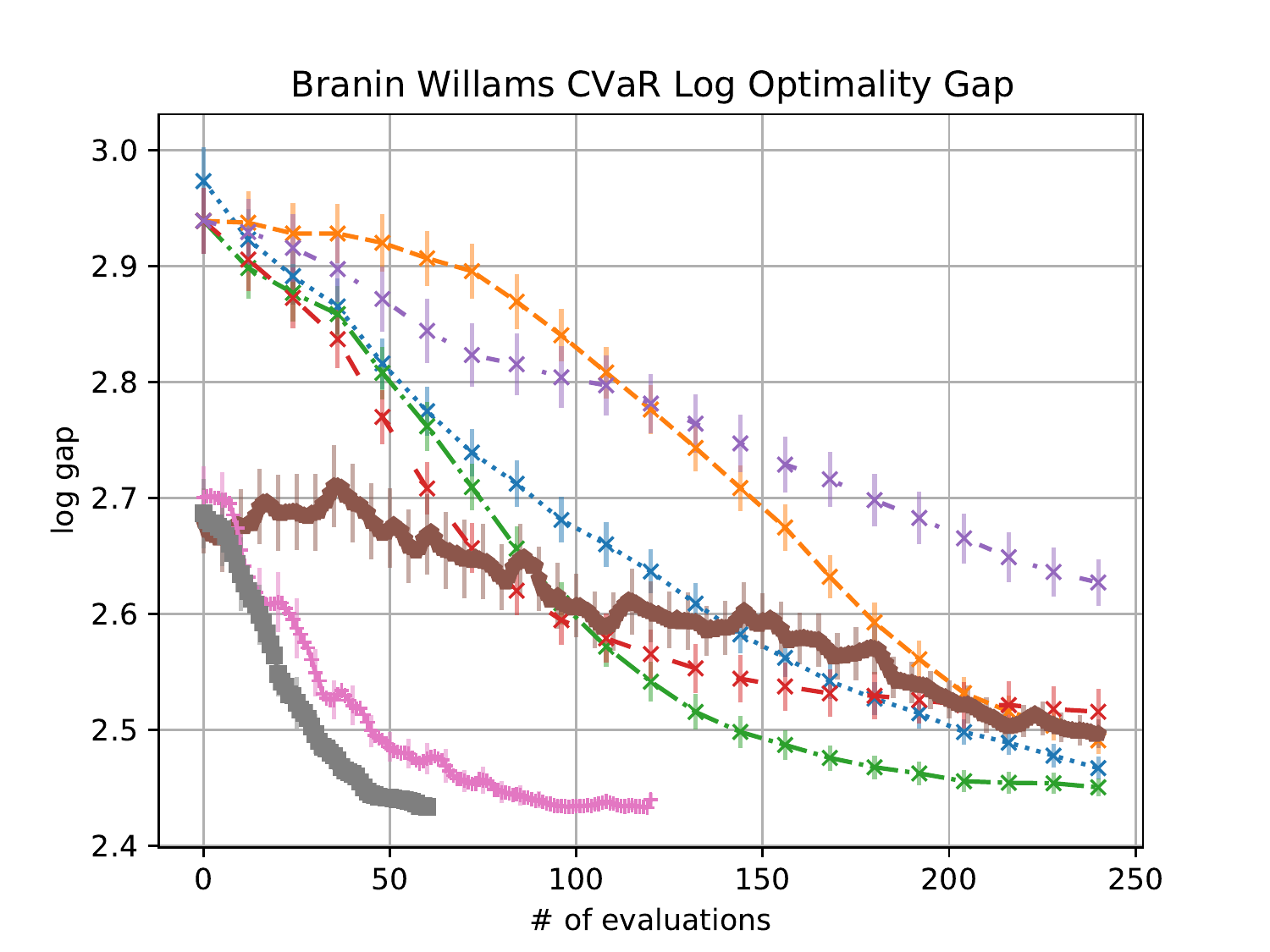} & \includegraphics[width=\linewidth]{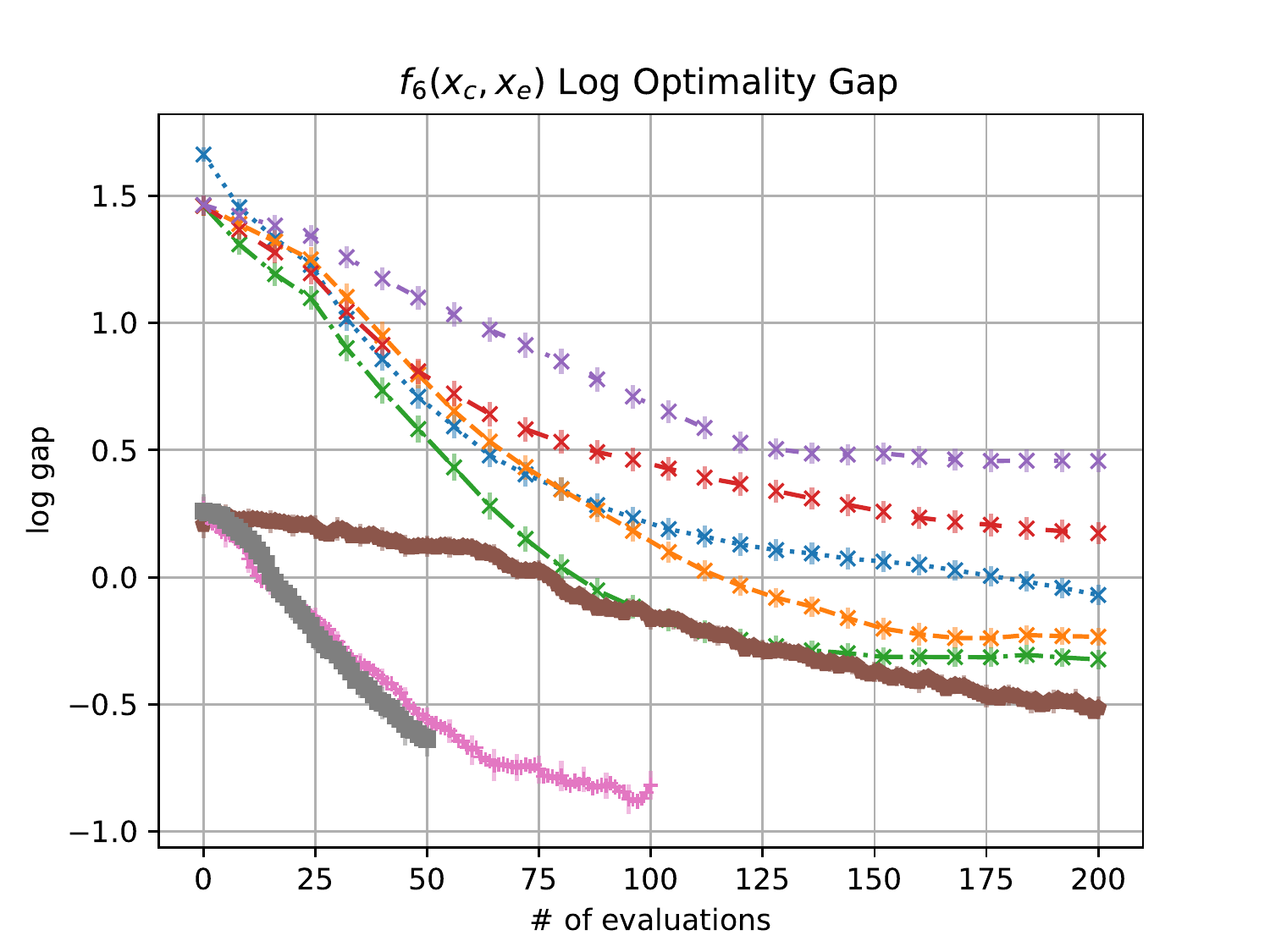} \\
        \includegraphics[width=\linewidth]{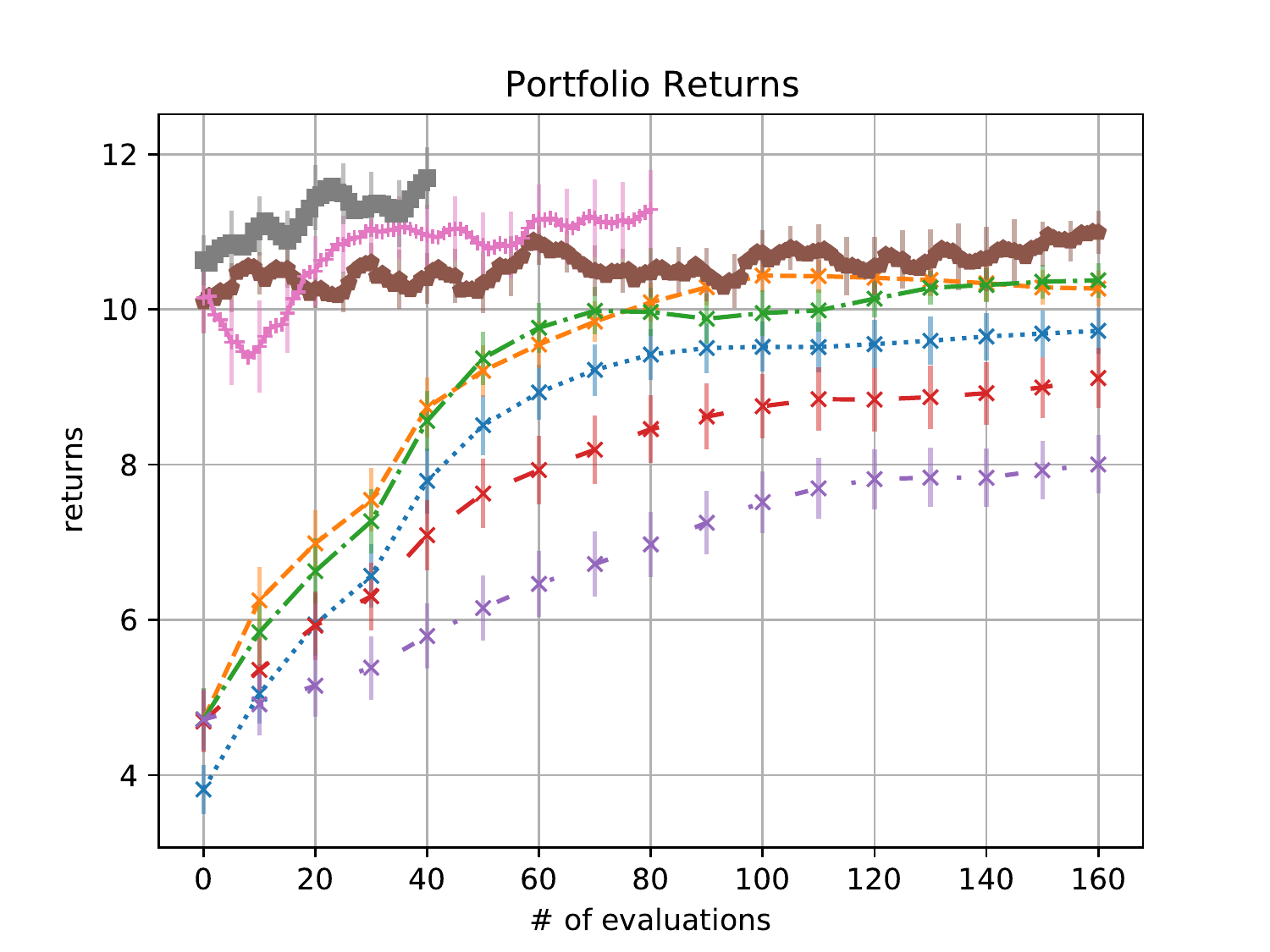} & \includegraphics[width=\linewidth]{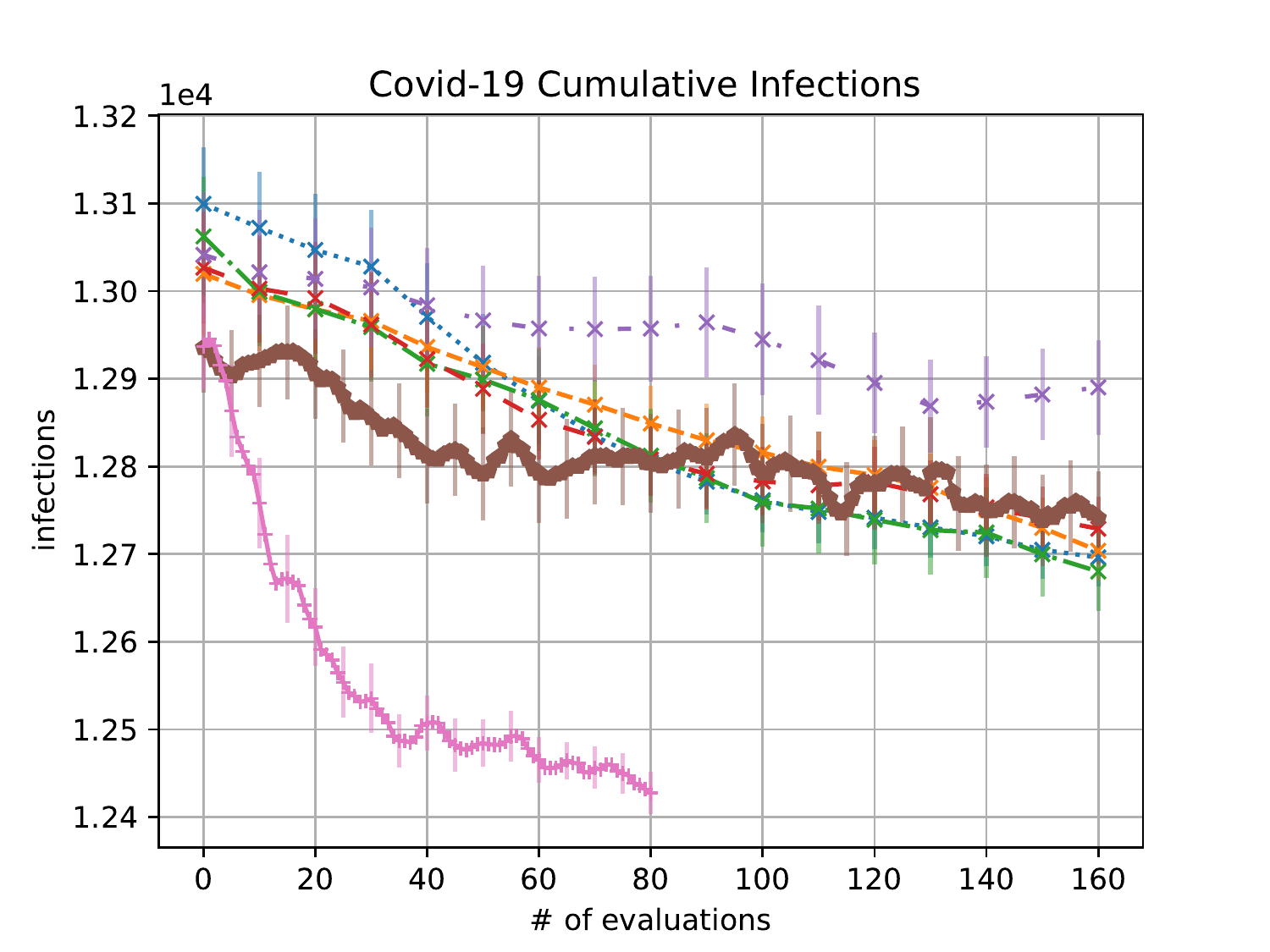} & \includegraphics[width=\linewidth]{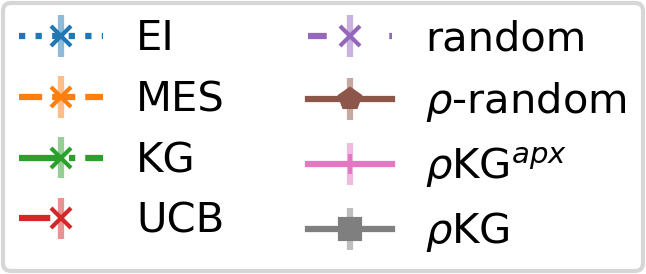}  \\
    \end{tabular}
    
    \caption{Top: The log optimality gap in Branin Williams with VaR (left) with CVaR (middle), and $f_6(x_c, x_e)$ (right). Bottom: The returns on Portfolio problem (left), the cumulative number of infections in the COVID-19 problem (middle) and the legend (right). The plots are plotted against the number of $F(x, w)$ evaluations, and are smoothed using a moving average of $3$ iterations. 
    Non-smoothed plots are given in the supplement.
    }
    \label{fig:exp-results}
\end{figure}

\subsection{Allocating COVID-19 testing capacity}

In this example, we study allocation of a joint COVID-19 testing capacity between three neighboring populations
(e.g., cities, counties). 
The objective is to allocate the testing capacity
between the populations 
to minimize the total number of people infected with COVID-19 in a span of two weeks. 
We use the COVID-19 simulator provided in 
\cite{covid-github} and discussed in \cite{CovidWhitepaper,forbes}.
The simulator models the interactions
between individuals 
within the population, the disease spread and progression, testing and quarantining of positive cases. Contact tracing is performed for people who test positive, and the contacts that are identified are placed in quarantine.

We study three populations of sizes $5\times 10^4$, $7.5 \times 10^4$, and $10^5$ that share a combined testing capacity of $10^4$ tests per day. The initial disease prevalence within each population is estimated to be in the range of $0.1-0.4\%, 0.2-0.6\%$ and $0.2-0.8\%$ respectively. We assign a probability of $0.5$ to the middle of the range and $0.25$ to the two extremes, independently for each population. Thus, the initial prevalence within each population defines the environmental variables. We pick the fraction of testing capacity allocated to the first two populations as the decision variable (remaining capacity is allocated to the third), with the corresponding decision space $\X = \{x \in \R^2_+: x_1 + x_2 \leq 1\}$. For the inner computations of $\acqfapx$, we use the full $\W$ set, however, for the evaluations of benchmark algorithms we randomly sample a subset $\widetilde{\W}$ of size $10$ to avoid using $27$ evaluations per iteration.

\subsection{Results}

Figure \ref{fig:exp-results}
plots results of the experiments.
Evaluations reported exclude the GP initialization, and the error bars denote one standard error. In each experiment, $\acqf$ and $\acqfapx$ match or beat the performance of all benchmarks using less than half as many function evaluations, thus, demonstrating superior sampling efficiency.
In the experiments $\widetilde{W}$ was intentionally kept small (between 8-12) to avoid giving our methods an outsized advantage. If $\widetilde{W}$ were larger, the benchmarks would use up even more evaluations per iteration, and our algorithms would provide an even larger benefit.

To demonstrate the benefit of using our novel statistical model, we included experiments with two random sampling strategies. The one labeled ``random'' evaluates $\rho[F(x, W)]$, and uses the corresponding GP model over $\X$. ``$\rho$-random'', on the other hand, evaluates $F(x,w)$ at a randomly selected $(x, w)$, uses our statistical model, and reports $\arg \min_x \E_n[\rho[F(x, W)]]$ as the solution. 
The ability to survey the whole $\X \times \W$ space gives ``$\rho$-random" a significant boost. We see that, despite choosing evaluations randomly, it is highly competitive against all the benchmarks, and outperforms ``random" by a significant margin. This demonstrates the added value of our statistical model, which captures all the information available in the data, and suggests an additional cheap-to-implement algorithm that is useful whenever $F(x, w)$ is cheap enough to render other algorithms too expensive.

The supplement presents additional plots comparing algorithm run-times. Data from Branin Williams and $f_6(x_c, x_w)$ show that even with only moderately expensive function evaluations (a few minutes per evaluation), our algorithms save time compared with the benchmarks presented here.

\section{Conclusion}\label{section-conclusion}
In this work, 
we introduced a novel Bayesian optimization approach for solving problems of the form $\min_x \rho[F(x, W)]$, where $\rho$ is a risk measure and $F$ is a black-box function that can be evaluated for any $(x, w) \in \X \times \W$. By modeling $F$  with a GP model instead of the objective function directly as is typical in Bayesian optimization, our approach is able to leverage more fine-grained information and, importantly, to jointly select both $x$ and $w$ at which to evaluate $F$. This allows our algorithms to significantly improve sampling efficiency over existing Bayesian optimization methods.

We propose two acquisition functions, $\acqf$, which is one step-Bayes optimal, and a principled cheap approximation, $\acqfapx$, along with an efficient, gradient-based approach to optimize them. To further improve numerical efficiency, we introduced a two time scale optimization approach that is broadly applicable for acquisition functions that require a nested optimization.

\section*{Broader Impact}

Our work is of interest whenever one needs to make a decision guided by an expensive simulator, and subject to environmental uncertainties. Such scenarios arise in simulation assisted medical decision making \cite{Jing2012Cardio}, in financial risk management \cite{Lopez1997Fed, McNeil2005, Gordy2010Nested}, in public policy, and disaster management \cite{Steward2007Disaster}.

The impact of our algorithms could be summarized as facilitating risk averse decision making. In many scenarios, risk averse approaches result in decisions that are more robust to environmental uncertainties compared to decisions resulting from common risk neutral alternatives. For example, in a financial crisis, an earlier risk-averse decision of holding more cash and other low-risk securities might prevent large losses or even the default of a financial institution. As another example, a risk averse decision of stockpiling of excess medical supplies in non-crisis times would alleviate the shortages faces during crisis times, such as the COVID-19 pandemic we are facing today.

On the negative side of things, the risk averse decisions we facilitate may not always benefit all stakeholders. For a commercial bank, a risk averse approach may suggest a higher credit score threshold for making loans, which might end up preventing certain groups from access to much needed credit.

In our case, the failure of the system would mean a poor optimization of the objective, and recommendation of a bad solution. Implementation of a bad decision can have harmful consequences in many settings; however, we imagine that any solution recommended by our algorithms would then be evaluated using the simulator, thus preventing the implementation of said decision.

Our methods do not rely on training data, and only require noisy evaluations of the function value. Thus, it can be said that our method does not leverage any bias in any training data. However, the solutions recommended by our algorithms are only good up to the supplied function evaluations, thus are directly affected by any biases built into the simulator used for these evaluations.

\section*{Acknowledgements}
The authors gratefully acknowledge the support by the National Science Foundation under Grants CAREER
CMMI-1453934 and CCF-1740822; and the Air Force Office of Scientific Research under Grants FA9550-19-1-0283 and FA9550-15-1-0038. We also thank the anonymous reviewers, whose comments helped improve and clarify the presentation of our paper.

\small

\bibliographystyle{ieeetr} 
\bibliography{ref.bib}

\end{document}

\maketitle

\tableofcontents

\section{Computational complexity}
% \begin{algorithm}[tb]
%   \caption{$\acqf$ Algorithm}
%   \label{alg:rhokg}
% \begin{algorithmic}
%   \STATE {\bfseries Input:} $N_0, N, M, K, L, \alpha$ 
%   \STATE Evaluate $F(\cdot, \cdot)$ at $N_0$ initial points.
% %   \STATE Initialize the GP using $N_0$ initial samples
%   \FOR{$n=1$ {\bfseries to} $N$}
%       \STATE Refit the GP hyper-parameters via MAP estimation.
%       \STATE Use the multi-start LBFGS algorithm to optimize VaRKG as described in Section \ref{optimization-section}, and set the maximizer $(x_{n+1}, w_{n+1})$ as the candidate to sample from.
%       \STATE Evaluate the candidate.
%   \ENDFOR
%   \STATE {\bfseries Return:} ${\arg \min}_{x} \E_N[ \rho[F(x, W)]]$ as the optimal solution.
% \end{algorithmic}
% \end{algorithm}

In this section, we analyze the computational complexity of using $\acqf$ to pick the $n$-th candidate to evaluate. 
For simplicity, we ignore the multiple restart points used for optimization, and present the analysis for a single restart point. The resulting complexity involves terms for number of LBFGS \cite{Zhu1997LBFGS} iterations performed. These terms can be adjusted to account for the cost of multiple restart points.

In this analysis, we assume that evaluating the GP prior mean and kernel functions (and the corresponding derivatives) takes $\mO(1)$ time.
We break down the computations into several parts and analyze them one by one. 

% Steps performed within an iteration are as follows:
% \begin{enumerate}
%     \item Refit the GP hyper-parameters via MAP estimation. %$\mO(Q S n^2 + Q n^3)$
%     \item Generate $K$ fantasy GP models.
%     % Use the precomputed Cholesky of $A_n$ to draw the fantasies - $\mO(S n + n^2 + K)$
%     % \item - the rest is per fantasy
%     % \item Compute the covariance matrix and its Cholesky factor
%     \item Use these to draw samples
%     \item Order the samples and compute $\rho$
% \end{enumerate}

An iteration starts by fitting the GP hyper-parameters using maximum a posteriori (MAP) estimation with $Q_1$ iterations of LBFGS algorithm. Each iteration requires calculating the likelihood of the data, which requires computing
\begin{equation*}
    A_n = \Sigma_0(x_{1:n}, w_{1:n}, x_{1:n}, w_{1:n}) + \text{diag}(\sigma^2(x_1, w_1), \ldots, \sigma^2(x_{n}, w_{n}))
\end{equation*} ($\mO(n^2)$ kernel evaluations), its Cholesky factor and inverse ($\mO(n^3)$), and solving a triangular set of equations ($\mO(n^2)$); putting the total cost of fitting the GP at $\mO(Q_1 n^3)$.

We can use the Cholesky factor of $A_n$ and the inverse $A_n^{-1}$ computed in the previous step to calculate the posterior mean and variance at the candidate point at a cost of $\mO(n^2)$, which are then used to draw $K$ fantasy observations at a cost of $\mO(K)$.

Let $A_{n+1}$ be the matrix obtained by adding $A_n$ one row and column corresponding to the candidate $(x_{n+1}, w_{n+1})$ at a cost of $\mO(n)$. Using the pre-computed Cholesky factor of $A_n$, we can compute the Cholesky factor of $A_{n+1}$ at a cost of $\mO(n^2)$. The inverse $A_{n+1}^{-1}$ can also be obtained from $A_n^{-1}$ at a cost of $\mO(n^2)$. Note that $A_{n+1}$ is identical for each fantasy model.

For each fantasy model, we need to compute the posterior mean and covariance matrix for the $L$ points $(x, w_{1:L})$, on which we draw the sample paths. 
Given $A_{n+1}^{-1}$, we first compute $\Sigma_0(x^i, w_{1:L}, x_{1:n+1}, w_{1:n+1}) A_{n+1}^{-1}$ at a cost of $\mO(L n^2)$, then use it to compute the mean vector and the covariance matrix at a total cost of $\mO(L^2 n)$. To sample from the multivariate normal distribution, we also need the Cholesky factor of the covariance matrix, which is obtained at a cost of $\mO(L^3)$.  Repeating this for each fantasy, we obtain all $K$ mean vectors, covariance matrices and the Cholesky factors at a total cost of $\mO(K L [n^2 +  L n + L^2])$.

Given the mean vector and the Cholesky factor of the covariance matrix, we can draw a sample of $F(x^i, w_{1:L})$ at a cost of $\mO(L^2)$. This results in a total cost of $\mO(K M L^2)$ to generate all samples. The sorting and other arithmetic operations needed for computing $\rho$ given the sample paths cost $\mO(L^2)$ per sample-path, thus introducing no additional terms.

Note that each iteration of LBFGS performs a bounded number of line search and a bounded number of function evaluations. 
Thus, if we use $Q_3$ iterations of LBFGS to solve the inner optimization problem, this translates to a total computational cost of 
$\mO(Q_3 K L [n^2 + L n + L^2 + M L])$
for the inner optimization of all $K$ fantasies.

If we then use $Q_2$ iterations of LBFGS to optimize the acquisition function, we end up with a total cost of 
$\mO(Q_2  Q_3 K L [n^2 + L n + L^2 + ML])$
for optimization, and a total cost of 
$\mO(Q_1 n^3 + Q_2  Q_3 K L [n^2 + L n + L^2 + ML])$ to pick the $n$-th candidate to evaluate, including the cost of fitting the GP model.

\section{The SAA problem}

In this work, we use the SAA approach \cite{Kleywegt2002SAA, Kim2015SAA} to optimize $\acqf$ and $\acqfapx$, and take advantage of the higher efficiency offered by deterministic optimization algorithms. The SAA approach trades a stochastic optimization problem with a deterministic approximation, which can be efficiently optimized. Suppose that we are interested in the optimization problem $\min_x \E_{\omega}[h(x, \omega)]$. The corresponding SAA problem is given by
\begin{equation*}
    \min_x \frac{1}{S} \sum_{i=1}^S h(x, \omega_i),
\end{equation*}
where $\omega_{1:S}$ are fixed realizations of $\omega$. It is well known that the optimal value of SAA lower bounds the original problem, and that it is consistent as $S\rightarrow \infty$.

\begin{algorithm}[tb]
  \caption{SAA of $\acqf$}
  \label{alg:SAA}
\begin{algorithmic}[1]
  \STATE {\bfseries Input:} The candidate $(x, w)$, the solutions to the inner problems $x^0_*$ and $x^i_*$, $K, M, L$.
  \STATE Fix the set $\widetilde{W} = w_{1:L}$, as well as the realizations of the random variables $Z_0^{i}$, $Z_L^{0j}$ and $Z_L^{ij}$, for $i=1:K, j = 1:M$.
  \STATE Use $Z_0^i$ to obtain the fantasy $\gp_f^i$ and its parameters $\mu_f^i, \Sigma_f^i$ as in equation (\ref{eq-fantasy-mu-sigma}).
  \STATE For sake of brevity, use $\gp_f^0, \mu_f^0, \Sigma_f^0$ to refer to the current $\gp, \mu_n, \Sigma_n$.
  \FOR{i = 0:K and j=1:M}
    \STATE Draw a sample of $F(x^i_*, w_{1:L})$ from $\gp_f^i$ as $\widehat{F}^{ij}(x^i_*, w_{1:L}) = \mu_f^i(x^i_*, w_{1:L}) + C_f^i Z_L^{ij}$ where $C_f^i$ is the Cholesky factor of $\Sigma_f^i(x^i_*, w_{1:L}, x^i_*, w_{1:L})$.
    \STATE Use the sample $\widehat{F}^{ij}(x^i_*, w_{1:L})$ with appropriate operations to get $\mathfrak{r}^{ij}(x^i_*)$, as explained in Section 4 of the paper. The corresponding gradient is then given by $\nabla_{(x, w)} \mathfrak{r}^{ij}(x^i_*)$.
  \ENDFOR
  \STATE Set $\widehat{\acqf}_n(x, w) = \frac{1}{M}\sum_{j=1}^M \mathfrak{r}^{0j}(x^0_*) - \frac{1}{KM}\sum_{i=1}^K \sum_{j=1}^M \mathfrak{r}^{ij}(x^i_*)$.
  \STATE Set $\nabla_{(x, w)} \widehat{\acqf}_n(x, w) = - \frac{1}{KM}\sum_{i=1}^K \sum_{j=1}^M \nabla_{(x, w)} \mathfrak{r}^{ij}(x^i_*)$.
  \STATE {\bfseries Return:} $\widehat{\acqf}_n(x, w)$ as the acquisition function value, and $\nabla_{(x, w)} \widehat{\acqf}_n(x, w)$ as the corresponding gradient.
\end{algorithmic}
\end{algorithm}

In the case of $\acqf$, the SAA problem cannot be written out explicitly, due to non-linearity of risk measures. However, the SAA problem and the corresponding gradient can still be obtained following an algorithmic approach, as explained in Algorithm \ref{alg:SAA}. The key to obtaining the SAA problem is using the reparameterization trick \cite{wilson2018maximizing} to write out the complex multivariate normal random variables as deterministic functions of a standard normal random vector, and the mean and covariance functions of the corresponding $\gp$ model. 

Let $Z_0^i, Z_L^{ij}$ be fixed realizations of the standard normal random vectors of dimension $1$ and $L$ respectively. The fantasy GP model, $\gp_f^i$, can be obtained as a continuous function of $(x, w)$ \cite{Wu2016qKG} as
\begin{equation} \label{eq-fantasy-mu-sigma}
    \begin{aligned}
        \mu_f^i(x', w') &= \mu_n(x', w') + \frac{\Sigma_n(x', w', x, w)}{\sqrt{\Sigma_n(x, w, x, w) + \sigma^2(x, w)}} Z_0^i, \\
        \Sigma_f^i(x', w', x'', w'') &= \Sigma_n(x', w', x'', w'') - \frac{\Sigma_n(x', w', x, w) \Sigma_n(x, w, x'', w'')}{\Sigma_n(x, w, x, w) + \sigma^2(x, w)}.
    \end{aligned}
\end{equation}
For positive definite matrices, the Cholesky factor is unique, and can be viewed as a differentiable function of the matrix \cite{murray2016differentiation}. In light of this fact, we can view the sample
\begin{equation*}
    \widehat{F}^{ij}(x^i, w_{1:L}) = \mu_f^i(x^i, w_{1:L}) + C_f^i Z_L^{ij}
\end{equation*}
as a deterministic and continuous function of both $(x, w)$ and $x^i$. For fixed $(x, w)$, $x^i_*$ can then be obtained, again using a SAA approach, by solving
\begin{equation}
    x^i_* = {\arg\min}_{x^i} \frac{1}{M}\sum_{j=1}^M \mathfrak{r}^{ij} (x^i);
\end{equation}
which is then used to obtain the SAA problem for $\acqf$:
\begin{equation*}
    \widehat{\acqf}_n(x, w) = \frac{1}{M}\sum_{j=1}^M \mathfrak{r}^{0j}(x^0_*) - \frac{1}{KM}\sum_{i=1}^K \sum_{j=1}^M \mathfrak{r}^{ij}(x^i_*).
\end{equation*}
The only possible gap left in the continuity / differentiability of $\widehat{\acqf}$ is the continuity of $\mathfrak{r}^{ij}(x^i)$ as a function of $\widehat{F}^{ij}(x^i, w_{1:L})$. This is conceptually identical to the continuity and almost sure differentiability of maximum of a finite number of continuous functions, where the maximizer only changes in the case of equality. In equation 3 of the paper and the preceding ordering, the ordering changes only in case of equality, which is a probability zero event. Thus, the resulting $\mathfrak{r}^{ij}(x^i)$ is differentiable with probability one. 
% We defer to Section \ref{sec-proofs} for a technical discussion on differentiability of $\mathfrak{r}^{ij}$ and the corresponding gradients of $\acqf$.

\section{Gradients of $\mathfrak{r}^{ij}$}
In this section, we make explicit the definition of the gradient $\nabla_{x^i} \mathfrak{r}^{ij}(x^i)$. In Section 4 of the paper, for a generic sample $\widehat{F}(x, w_{1:L})$, we define
\begin{equation*}
    \widehat{v}(x) := \widehat{F}(x, w_{(\lceil L \alpha \rceil)}) \quad \text{ and } \quad  \widehat{c}(x) := \frac{1}{\lceil L (1 - \alpha) \rceil} \sum_{j = \lceil L \alpha \rceil}^L \widehat{F}(x, w_{(j)}),
\end{equation*}
as the corresponding MC samples of VaR and CVaR respectively. For $\mathfrak{r}^{ij}$, using the reparameterization trick, we can write the corresponding samples of $F$ as,
\begin{equation*}
    \widehat{F}^{ij}(x^i, w_{1:L}) = \mu_f^i(x^i, w_{1:L}) + C_f^i Z_L^{ij},
\end{equation*}
and the corresponding $\mathfrak{r}^{ij}(x^i)$ as
\begin{equation*}
    \widehat{v}^{ij}(x^i) := \widehat{F}^{ij}(x^i, w_{(\lceil L \alpha \rceil)}) \quad \text{ and } \quad  \widehat{c}^{ij}(x) := \frac{1}{\lceil L (1 - \alpha) \rceil} \sum_{j = \lceil L \alpha \rceil}^L \widehat{F}^{ij}(x^i, w_{(j)}),
\end{equation*}
where the ordering of $w_{(\cdot)}$ depends on the particular realization of $\widehat{F}^{ij}(\cdot, \cdot)$ as well as the value of $x^i$. For fixed base samples $Z_L^{ij}$, this is a deterministic function of $\mu_f^i, C_f^i$ and $x^i$; where $\mu_f^i, C_f^i$ are differentiable in $x^i$, as well as $(x, w)$. Keeping the ordering of $w_{(\cdot)}$, the gradient can then be written as 
\begin{equation*}
    \nabla_{x^i} \widehat{v}^{ij}(x^i) := \nabla_{x^i} \widehat{F}^{ij}(x^i, w_{(\lceil L \alpha \rceil)}) \quad \text{ and } \quad \nabla_{x^i} \widehat{c}^{ij}(x) := \frac{1}{\lceil L (1 - \alpha) \rceil} \sum_{j = \lceil L \alpha \rceil}^L \nabla_{x^i} \widehat{F}^{ij}(x^i, w_{(j)}),
\end{equation*}
where $\nabla_{x^i} \widehat{F}^{ij}(x^i, w_{(\cdot)})$ is explicitly given by
\begin{equation*}
    \nabla_{x^i} \widehat{F}^{ij}(x^i, w_{(\cdot)}) = \nabla_{x^i} \mu_f^i(x^i, w_{1:L}) + \nabla_{x^i} C_f^i Z_L^{ij}.
\end{equation*}
The treatment of gradient $\nabla_{(x, w)} \mathfrak{r}^{ij}(x^i)$ is identical to the one presented here. Additional discussion on the theoretical properties of these gradients can be found in Section \ref{sec-proofs} of this supplement. 

\section{Two time scale optimization}

Before starting the discussion on the TTS approach, it is helpful to define the term ``raw samples". In many settings, as well as in this paper, non-convex functions are optimized using a multi-start gradient-based approach. Multi-start optimization requires selecting a set of restart points, which are typically randomly drawn from the solution space in the absence of domain-specific knowledge. A gradient-based algorithm then performs optimization starting from each of these restart points, which has a cost that is linear in the number of restart points. Raw samples offer an alternative way of selecting these restart points: Consider drawing a large number of points randomly from the solution space; then, instead of performing gradient-based optimization starting from each of these points, consider evaluating the objective function once for each point and then selecting a subset of them from which to start gradient-based optimization using a softmax approach.  We call this larger number of randomly selected points {\it raw samples} and we refer to the points selected for gradient-based optimization as {\it restart points}.
This approach is better able to select high-quality restart from which to perform gradient-based optimization without the need for domain-specific knowledge. This enables us to use a smaller number of restart points for optimization while still providing a high-quality global solution. 

Among state-of-the-art BO algorithms, knowledge gradient (KG) algorithms are on the more expensive side due to their underlying nested optimization structure. Although the added computational cost is typically justified by the superior sampling efficiency they offer (see \cite{Wu2016qKG,balandat2019botorch} for numerical comparisons), reducing the computational cost would make KG algorithms applicable in much broader settings. 

To reduce the computational cost of KG optimization, \cite{balandat2019botorch} proposes a one-shot optimization approach, which converts the $d^\X$ dimensional nested optimization problem of the KG acquisition function \cite{Wu2016qKG},
\begin{equation*}
    \max_x \frac{1}{K} \sum_{i=1}^K \max_{x^i} \mu_f^i(x^i),
\end{equation*}
into a single $(K + 1) d^\X$ dimensional optimization problem:
\begin{equation*}
    \max_{x, x^i, i=1, \ldots, K} \frac{1}{K} \sum_{i=1}^K \mu_f^i(x^i).
\end{equation*}
The objective function of the one-shot problem is substantially cheaper to evaluate, since it does not require solving an optimization problem. Moreover, this new problem is deterministic and smooth, allowing the use of efficient gradient-based algorithms such as LBFGS. However, the one-shot problem is again non-convex, and its large dimension makes it challenging to solve to global optimality.

Using a gradient-based approach, the one-shot problem is optimized locally from a given set of restart points, resulting in a locally optimal solution. 
In particular, the choice of some of the $x^i$ can be locally optimal, in contrast with the globally optimal value for $x^i$ found by a nested optimization approach with sufficiently many restarts.
% to the inner \pfedit{optimization} problems {\pfedit over $x^i$}, rather than a global solution one would get with a nested optimization approach. 
If $K$ is large, and the majority of the inner solutions are at their global optimum, the solutions that are not globally optimized do not significantly affect the algorithm's performance. However, getting the majority at their global optimum requires starting with a large number of raw samples.

In the classical BO setting, because evaluations of $\mu^i_f(x^i)$ are cheap, one can afford to evaluate the one-shot acquisition function at a large number of (carefully crafted) raw samples, and select a good set of restart points for optimization. As $K$ increases, both the number of raw samples and the restart points need to increase as well. 
When the acquisition function evaluations (even without the inner optimization) are significantly more expensive than in the classical setting (e.g., $\acqf$ requires evaluating $\frac{1}{M}\sum_{j=1}^M \mathfrak{r}^{ij}(x^i)$), one is restricted to a relatively small number of raw samples, which requires using a small $K$ ($\approx 10$) to achieve a good performance within a reasonable time. 

Consider extending this approach to our setting, where we jointly maximize over candidates $(x,w)$ and $x^i, i=1,\ldots,K$, and $\mu^i_f(x^i)$ is replaced by the posterior mean of $\rho[F(x, W)]$.
Since this posterior mean is more expensive to evaluate than
$\mu^i_f(x^i)$, it becomes  computationally burdensome to use a large number of raw samples.
Using few raw samples, unfortunately, can also cause problems:
a good candidate $(x, w)$ can appear poor because many of its $x^i$ are at local optima; 
while a poor candidate $(x,w)$ can outperform it because more of its $x^i$ happened to reach globally optimal values.
In such a scenario, the one-shot approach will recommend the bad candidate with the large acquisition function value, leading to a poor sampling decision. This so-called {\it instability} became an issue with the one-shot implementation of $\acqf$ in our preliminary testing and led to a search for a more {\it stable} alternative.

In this paper, we propose {\it two time scale optimization} (TTS), a novel approach for optimizing KG type of acquisition functions that addresses the issues raised here about the one-shot approach while still providing significant computational savings. 
The main idea behind the TTS approach is that the objective of the inner optimization problem is a continuous function of the candidate; and the optimal inner solution changes only a small amount between iterations when we update the candidate using a gradient-based algorithm. Thus, the solution to the inner problem obtained with the previous candidate provides a high quality approximation of the acquisition function value and its gradient. TTS optimization leverages this observation to provide computational savings by solving the inner optimization problem once every $T$ iterations, and reusing the obtained solution in between.
We provide a detailed description of TTS in Algorithm \ref{alg:tts}. We note that LBFGS performs multiple line searches within an iteration, and evaluates the solution found at the end of each line search. The description of Algorithm \ref{alg:tts} treats each line search as an iteration of LBFGS, and the candidate $(x(t), w(t))$ is updated at the end of each line search. This essentially means that we optimize the inner solution at every $T$-th evaluation of the $\acqf$ and not necessarily at every $T$-th iteration of LBFGS.

\begin{algorithm}[tb]
  \caption{Two time scale optimization of $\acqf$}
  \label{alg:tts}
\begin{algorithmic}
  \STATE {\bfseries Input:} $Q_2, Q_3, M, K, L, \alpha$, starting candidate $(x(0), w(0))$, and TTS-frequency $T$.
  \FOR{$t=0$ {\bfseries to} $Q_2-1$}
    \STATE Generate the fantasy models $\gp_f^i, i=1, \ldots, K$ using the candidate $(x(t), w(t))$.
    \FOR{$i=1$ {\bfseries to} $K$}
      \IF{$t \bmod T = 0$}
        \STATE Generate a set of random restart points for optimization of $i$-th inner problem. Include the previous solution $x^i(T)$ if available.
        \STATE Use $Q_3$ iterations of the multi-start LBFGS algorithm to optimize $i$-th inner problem, and set $x^i(T)$ as the optimizer.
      \ENDIF
      \STATE Return $x^i(T)$ as the solution to $i$-th inner problem
    \ENDFOR
    \STATE Set $\frac{1}{K} \sum_{i=1}^K - \frac{1}{M} \sum_{j=1}^M \nabla_{(x, w)} \mathfrak{r}^{ij}(x^i(T))$ as the gradient at the current candidate.
    \STATE Do an iteration of LBFGS using this gradient to obtain the new candidate $(x(t+1), w(t+1))$.
  \ENDFOR
  \STATE {\bfseries Return:} $(x(Q_2), w(Q_2))$ as the next candidate to evaluate.
\end{algorithmic}
\end{algorithm}

In the algorithm description, the parameter $T$ is the {\it TTS-frequency} that determines how often the inner problem is solved. If $T=1$, the TTS optimization reduces to the standard nested optimization. In numerical testing, we observed that the TTS optimization reduces the cost of optimizing $\acqf$ roughly by a factor of $T$. In our experiments, we used $T=10$ without any noticeable change in algorithm performance. We emphasize that TTS optimization can be used with other BO algorithms that involve a nested optimization, including all knowledge gradient type acquisition functions.

We mentioned earlier that the TTS optimization would address the {\it instability} we faced with the one-shot optimization. Going back to our example with one good and one bad candidate, TTS, by periodically solving the inner problem globally for each candidate, ensures that each candidates acquisition function value is calculated using a global solution, reducing the chances of misidentifying the bad candidate as the better one. 

We conclude by noting that the TTS implementation presented here is a primitive one, with room for improvement. We would also like to point out that our discussion of one-shot optimization is mostly an observational one, based on certain shortcomings we faced. An in-depth comparison and further development of the two optimization methods is left as a subject for future research.

\section{Experiments}
\subsection{Algorithm parameters}
\begin{itemize}[leftmargin=*]
    \item Common parameters: Each algorithm is optimized using the LBFGS algorithm \cite{Zhu1997LBFGS} (except for Covid-19 where we use SLSQP \cite{Kraft1988SLSQP} due to the linear constraint on the decision variables), with $10 \times (d^\X + d^\W)$ random restart points which are selected from $500 \times (d^\X + d^\W)$ raw samples. We use low-discrepancy Sobol sequences \cite{OWEN1998Sobol} to generate the raw samples. The restart points are selected using a softmax heuristic implemented in the BoTorch package \cite{balandat2019botorch}, which assigns each raw sample a weight that is an exponential factor of its acquisition function value, then selects the restart points with probability proportional to this weight.
    
    \item EI: Does not have any specific parameters.
    
    \item MES: A set of $500 \times (d^\X + d^\W)$ randomly drawn points is used to discretize the design space, and the max values are sampled over this discretization using the Gumbel approximation.
    We draw $10$ max value samples over the discretization, and use $128$ fantasies to compute the approximate information gain from using the candidate.
    
    \item KG: We use the one-shot KG implementation with $64$ fantasies.
    
    \item UCB: We use parameter $\beta = 0.2$, i.e. the acquisition function is given by $\text{UCB}(x) = \mu_n(x) + \sqrt{0.2 \Sigma_n(x, x)}$.
    
    \item $\acqf$ and $\acqfapx$: We use $K = 10$ fantasy models for optimization and $K = 4$ fantasy models to evaluate the raw samples; and generate $M = 10$ sample paths per fantasy model. For the inner optimization problem of $\acqf$, we use $5 \times d^\X$ random restart points selected from $50 \times d^\X$ raw samples. We use TTS optimization with $T=10$.
\end{itemize}

All fantasy models and sample paths are generated using low-discrepancy Sobol sequences \cite{OWEN1998Sobol}. For our algorithms, we observed a small improvement when increasing the number of fantasy models, however, this comes with a linear increase in run time and memory usage. We used a smaller number of fantasy models to evaluate the raw samples, as this reduces the computational cost without affecting performance. The number of sample paths were chosen rather arbitrarily, since the majority of computational effort is spent in calculating the Cholesky factor, which is then used for generating all sample paths. We observed very similar performance with $M = 4, 10$ and $40$; and decided to use $M=10$ rather arbitrarily.

\subsection{GP model and fitting}
We use the ``SingleTaskGP'' model from the BoTorch package \cite{balandat2019botorch} with the given priors on hyper-parameters, which, quoting from the code, ``... work best when covariates are normalized to the unit cube and outcomes are standardized (zero mean, unit variance)''. We use the Mat\'ern 5/2 kernel and fit the hyper-parameters using MAP estimation. 

We project the function domain $\X \times \W$ to the unit hypercube $[0,1]^{d^\X + d^\W}$, and all the calculations are done on the unit hypercube. For calculations involving the GP, we use the ``Standardize'' transformation available in BoTorch, which transforms the outcomes (observations) into zero mean, unit variance observations for better compatibility with the hyper-parameter priors.

\subsection{Synthetic test problems}
\begin{figure}[ht]
    \centering
    \includegraphics[width=0.3\linewidth]{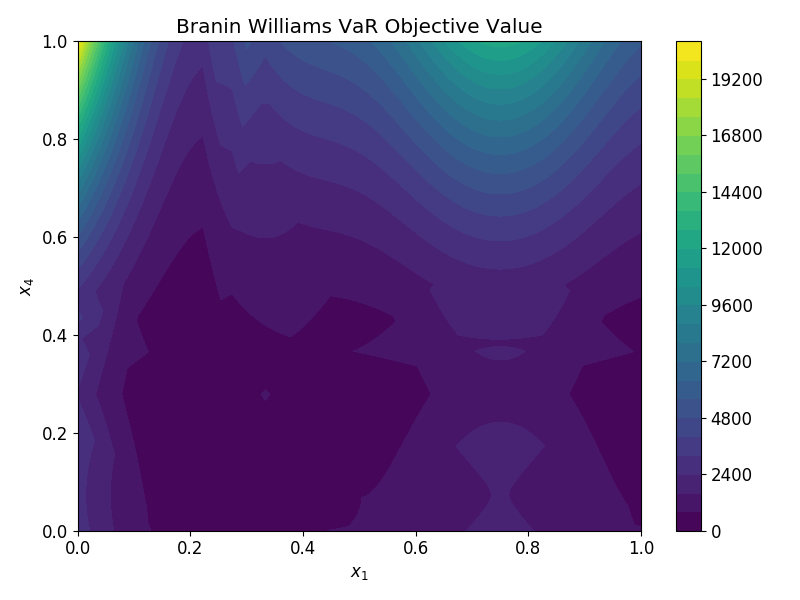}
    \includegraphics[width=0.3\linewidth]{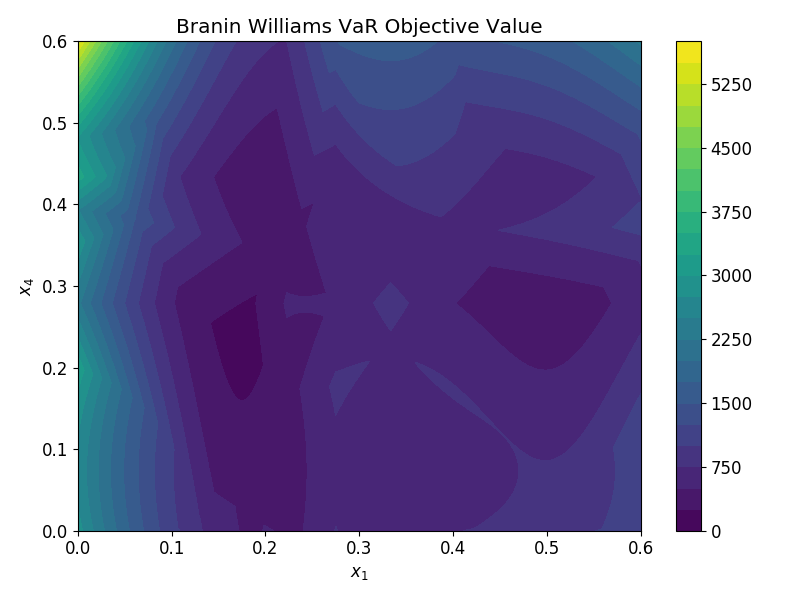}
    \includegraphics[width=0.3\linewidth]{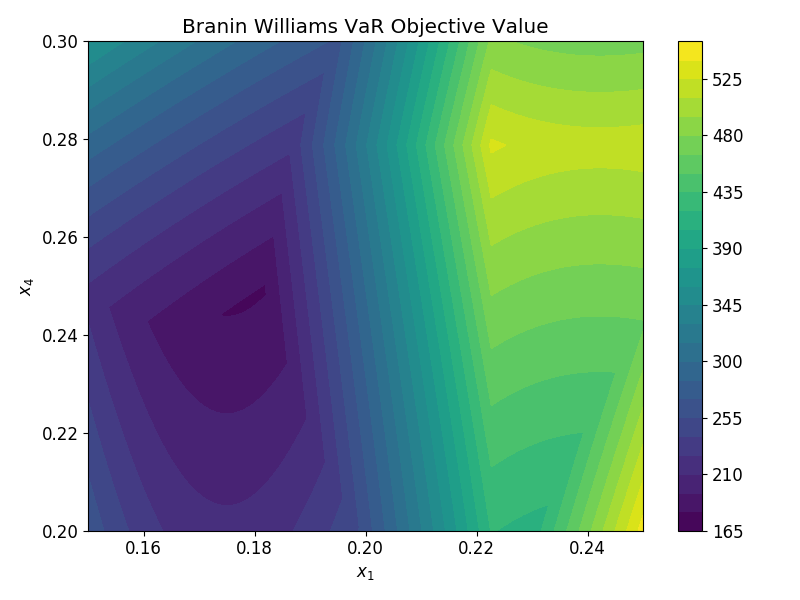}
    \caption{The VaR objective of Branin Williams function. Left to right, full domain and close-ups of the optimal region. It is notable that the objective function is quite flat across most of its domain, making it challenging to identify the optimal solution using a small number of evaluations.}
    \label{fig:bw-objective}
\end{figure}

The first two problems we use are synthetic test functions from the BO literature. The first problem is the 4 dimensional Branin-Williams problem formulated in \cite{Williams2000BW}. For $x \in [0,1]^4$, the function is defined as $y(x) = y_b(15 x_1 - 5, 15 x_2) \times y_b(15 x_3 - 5, 15 x_4)$, where
\begin{equation*}
    y_{b}(u, v)=\left(v-\frac{5.1}{4 \pi^{2}} u^{2}+\frac{5}{\pi} u-6\right)^{2}+10\left(1-\frac{1}{8 \pi}\right) \cos (u)+10.
\end{equation*}
The function is observed with additive Gaussian noise with a standard deviation of 10. The decision variables are $x_1$ and $x_4$, and the environmental variables are $w = (x_2, x_3)$. The set $\W$ of size $12$ and the corresponding probability distribution are given in Table \ref{tab:bw-distribution}.

The problem is originally formulated for the expectation objective, however, here we consider the minimization of VaR at risk level $\alpha = 0.7$. A plot of the VaR objective of the Branin Willams function is found in Figure \ref{fig:bw-objective}.

The second problem we consider is $f_6(x_c, x_e)$ function from \cite{marzat2013} given by
\begin{align*}
    f_{6}\left(x_{c}, x_{e}\right) &= x_{e 1}\left(x_{c 1}^{2}-x_{c 2}+x_{c 3}-x_{c 4}+2\right)+x_{e 2}\left(-x_{c 1}+2 x_{c 2}^{2}-x_{c 3}^{2}+2 x_{c 4}+1\right) \\
    &+ x_{e 3}\left(2 x_{c 1}-x_{c 2}+2 x_{c 3}-x_{c 4}^{2}+5\right)+5 x_{c 1}^{2}+4 x_{c 2}^{2}+3 x_{c 3}^{2}+2 x_{c 4}^{2}-\sum_{i=1}^{2} x_{e i}^{2},
\end{align*}
where $x_c \in [-5, 5]^4$ are the decision variables and $x_e \in [-2, 2]^3$ are the environmental variables. The function is evaluated with additive Gaussian noise with a standard deviation of $1$. We formulate this problem for minimization of CVaR at risk level $\alpha$ where $w = x_e$ has a continuous uniform distribution over its domain. For the computation of $\acqf$ and $\acqfapx$, we use a randomly drawn $\widetilde{\W}$ of size $40$ at each iteration, and use randomly drawn $\widetilde{\W}$ of size $8$ for evaluating $\text{CVaR}[F(x, W)]$ in benchmarks.

\subsection{Experiment details}
We initialize the GP model for the benchmark algorithms using $2 d^{\X} + 2$ random samples of $x_j \in \X$, and evaluate (or estimate) $\rho[F(x_j, W)]$ using evaluations of $F(x_j, w), w \in \widetilde{\W}$ where $\widetilde{\W}$ is the set corresponding to the benchmark algorithms as described below.
For $\acqf$ and $\acqfapx$, we use equivalent number of evaluations of $F(x, w)$, using $(x_j, w_j)$ randomly drawn from $\X \times \W$. Each replication of the experiments uses a new draw of the samples; and the samples are synchronized across different algorithms.

In what follows, we describe the use of $\W$ in each experiment. Note that all random sets $\widetilde{\W}$ are independently redrawn at each iteration.
\begin{enumerate}[leftmargin=*]
    \item Branin-Williams: The set $\W$ and the corresponding probability distribution is given in Table \ref{tab:bw-distribution}. Our algorithms use the full set $\W$ in the inner computations, and all the benchmarks evaluate $\text{VaR}_{0.7}[F(x, W)]$ using the full $\W$, thus using $12$ evaluations per iteration.
    
    \begin{table}
      \caption{Probability distribution of $x_2$ and $x_3$ in Branin Williams problem. }
      \label{tab:bw-distribution}
      \centering
        \begin{tabular}{cccccc}
            \toprule
             & & \multicolumn{4}{c} {$x_{3}$} \\
            % \cmidrule(r){3-6}
             & & 0.2 & 0.4 & 0.6 & 0.8 \\
             \midrule 
            \multirow{3}{*}{$x_2$} & 0.25 & 0.0375 & 0.0875 & 0.0875 & 0.0375 \\
             & 0.5 & 0.0750 & 0.1750 & 0.1750 & 0.0750 \\
             & 0.75 & 0.0375 & 0.0875 & 0.0875 & 0.0375 \\
            \bottomrule
        \end{tabular}
    \end{table}
    
    \item $f_6(x_c, x_e)$: The set $\W$ is the hyper-rectangle $[-2, 2]^3$ with a continuous uniform distribution. At each iteration of our algorithms, we sample a subset $\widetilde{\W}$ of size 40, and use this for the inner computations. The $w$ component of the candidate is not restricted to this subset $\widetilde{\W}$, and can take any value in $\W$. For each evaluation of the benchmarks, we sample a subset $\widetilde{\W}$ of size $8$, and calculate $\text{CVaR}_{0.75}[F(x, W)]$ on this set, thus using $8$ evaluations per iteration.
    
    \item Portfolio optimization: The set 
    \begin{equation*}
        \W = \{w \in \R^2: w_1 \in [10^{-4}, 10^{-2}], w_2 \in [10^{-4}, 10^{-3}]\}
    \end{equation*} gives the range of bid-ask spread ($w_1$) and the borrowing cost ($w_2$), and $W$ follows a uniform distribution on $\W$. For the inner computations of our algorithms, we use a randomly drawn subset $\widetilde{\W}$ of size $40$, and use a subset $\widetilde{\W}$ of size $10$ for benchmark evaluations of $\text{VaR}_{0.8}[F(x, w)]$, thus using $10$ evaluations per iteration.
    
    \item COVID-19: The set $\W$ of initial disease propensity is given by
    \begin{equation*}
        \W = \left \{w \in \R^3: \begin{array}{c}
             w_1 \in \{ 0.001, 0.0025, 0.004 \}, \\
             w_2 \in \{0.002, 0.004, 0.006 \}, \\
             w_3 \in \{0.002, 0.005, 0.008\}
        \end{array}  \right\}
    \end{equation*}
    with the probability distribution given by 
    \begin{equation*}
        \mathbb{P}_W(W = w) = 2^{-8 + \mathds{1}(w_1 = 0.0025) + \mathds{1}(w_2 = 0.004) + \mathds{1}(w_3 = 0.005)}.
    \end{equation*}
    For the inner computation of $\acqfapx$, we use the full set $\W$. For evaluations of the benchmarks, we draw a random sample $\widetilde{\W}$ of size $10$ and use this to calculate $\text{CVaR}_{0.9}[F(x, W)]$, thus using $10$ evaluations per iteration.
    
    The output of the Covid-19 simulator is stochastic with a large variation. To enable fair comparison between the algorithms, we fix a set of $10$ seeds, and each function evaluation is done using a randomly selected seed from this set. The reported performance is computed as the average performance of using all $10$ seeds.
\end{enumerate}

We ran the benchmark algorithms EI, MES, KG, UCB and random for $100$ replications in each experiment. For $\rho-$random and $\acqfapx$, we ran 50 replications in each experiment. Finally, $\acqf$ was ran for $30$ replications in each experiment, except for COVID-19.

\subsection{Results}
\begin{figure}[ht]
    \centering
    \begin{tabular}{m{0.30\linewidth} m{0.30\linewidth} m{0.30\linewidth} }
      \includegraphics[width=\linewidth]{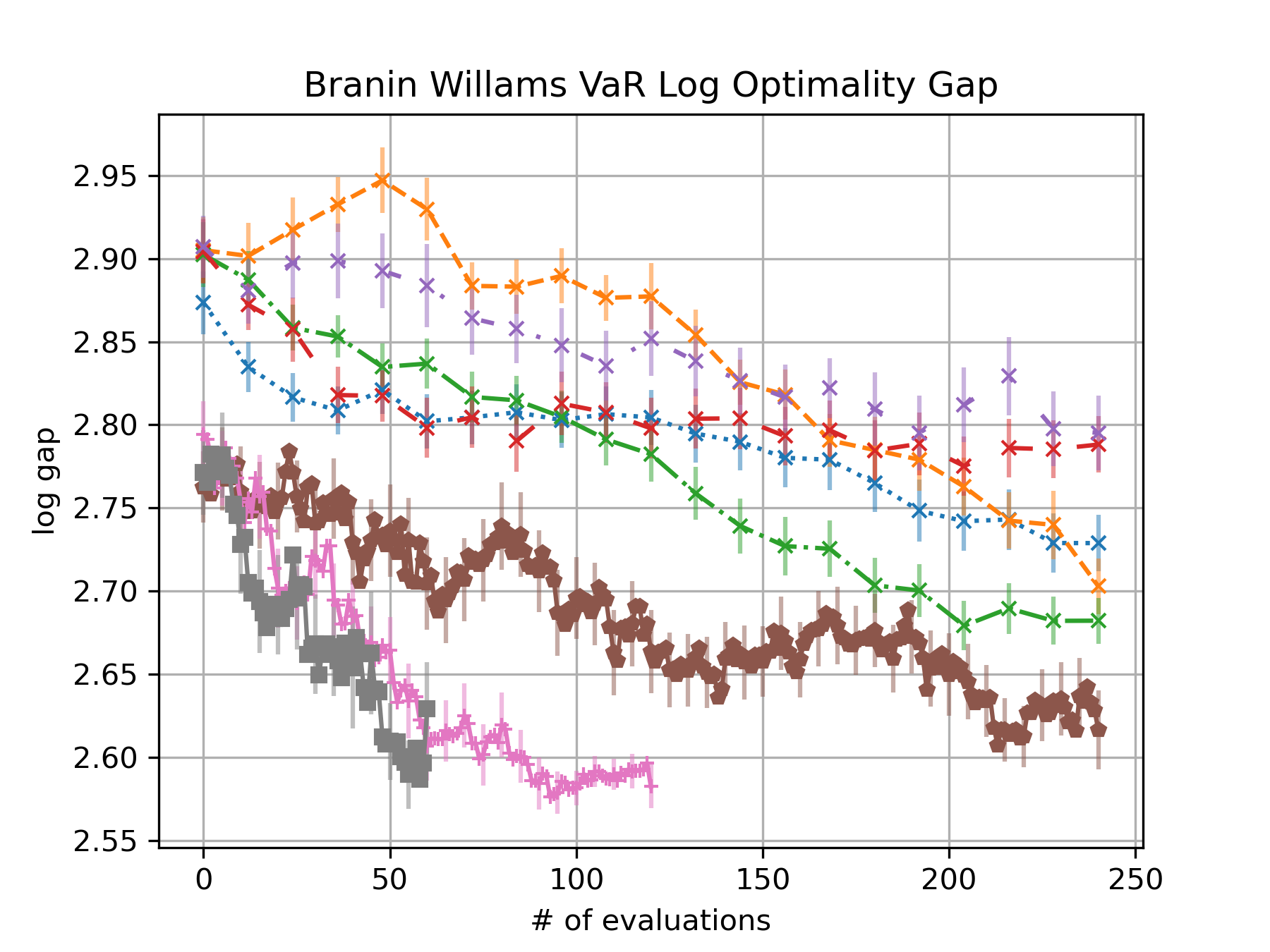}  & \includegraphics[width=\linewidth]{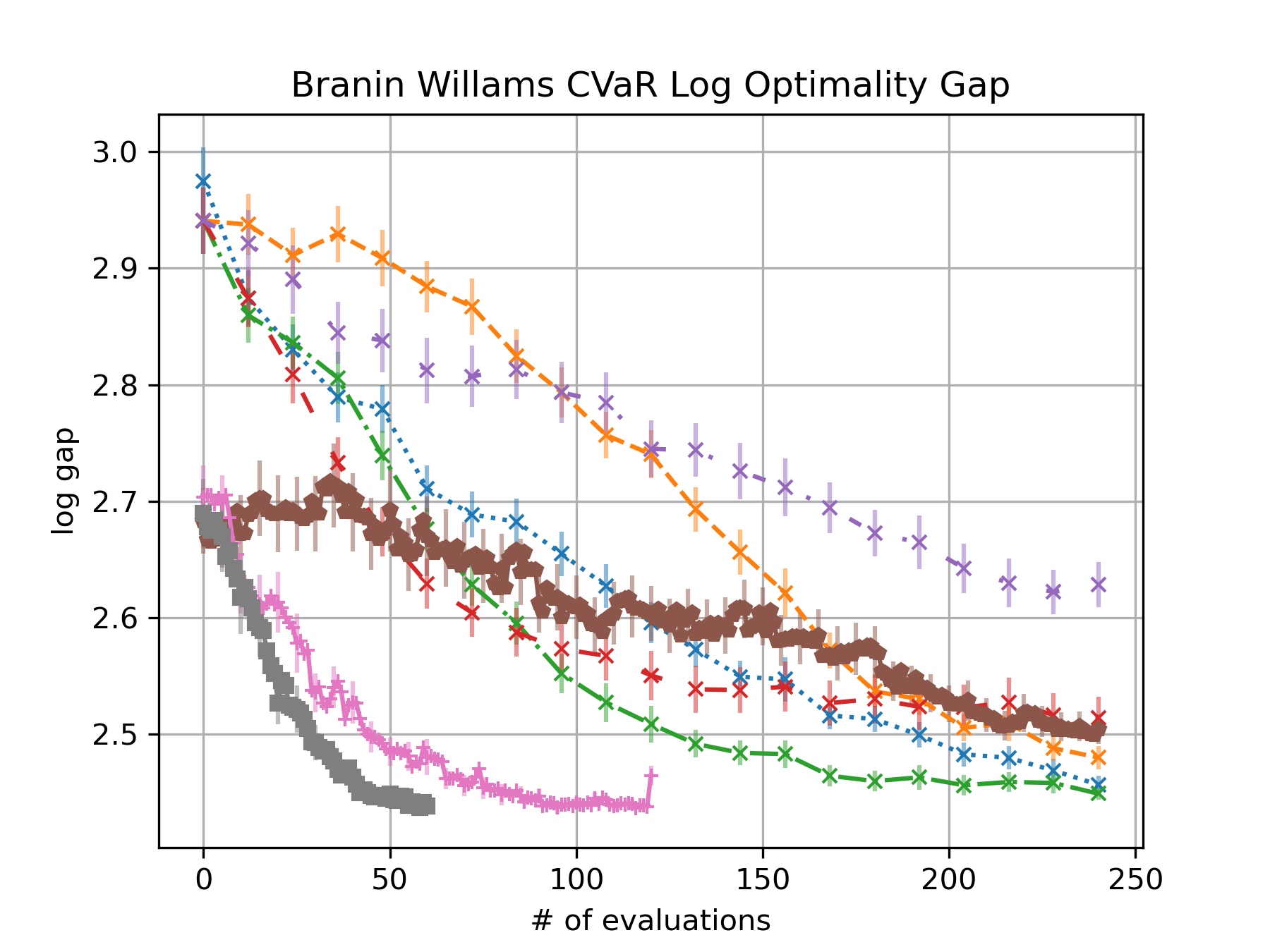} & \includegraphics[width=\linewidth]{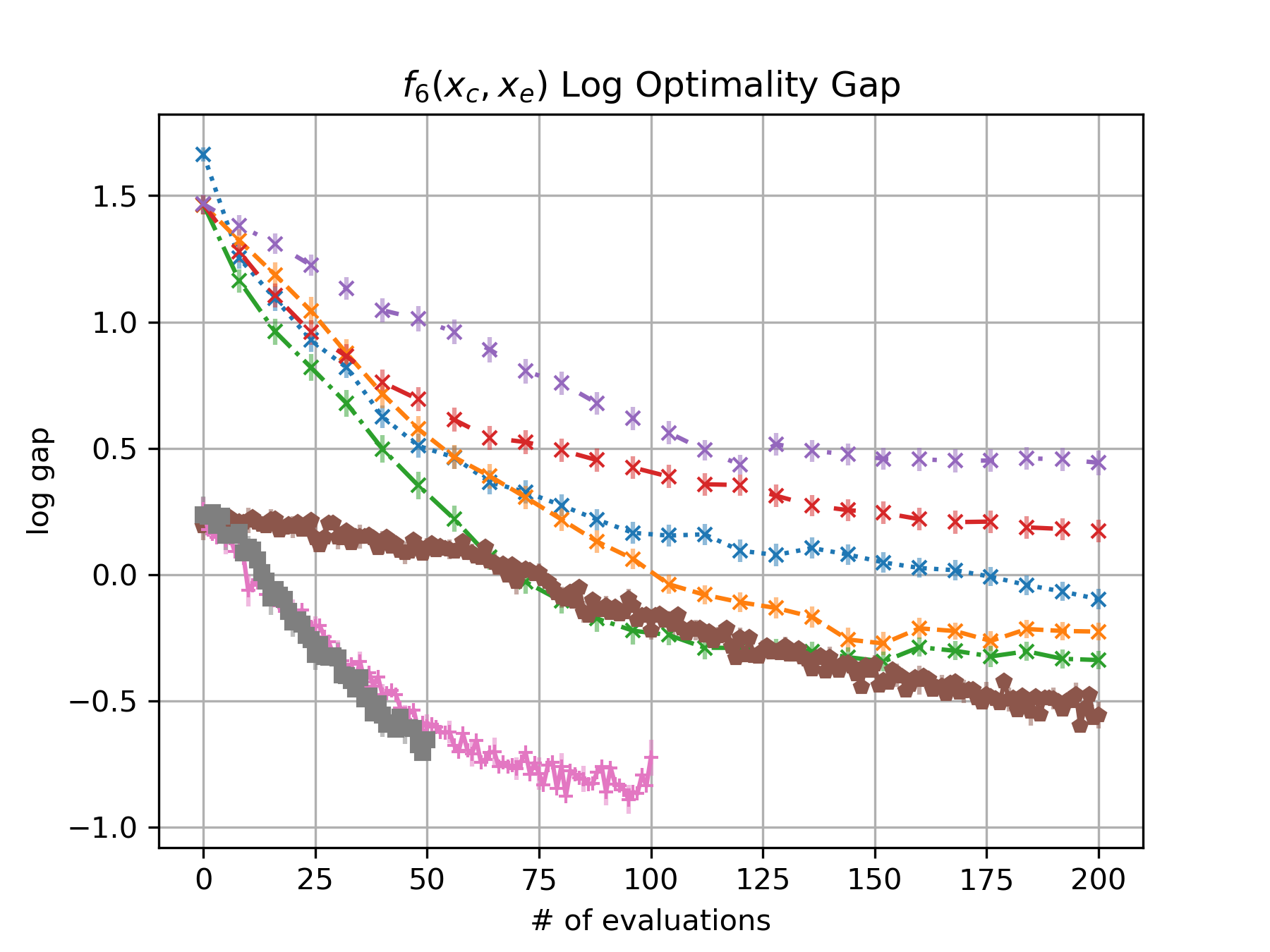} \\
        \includegraphics[width=\linewidth]{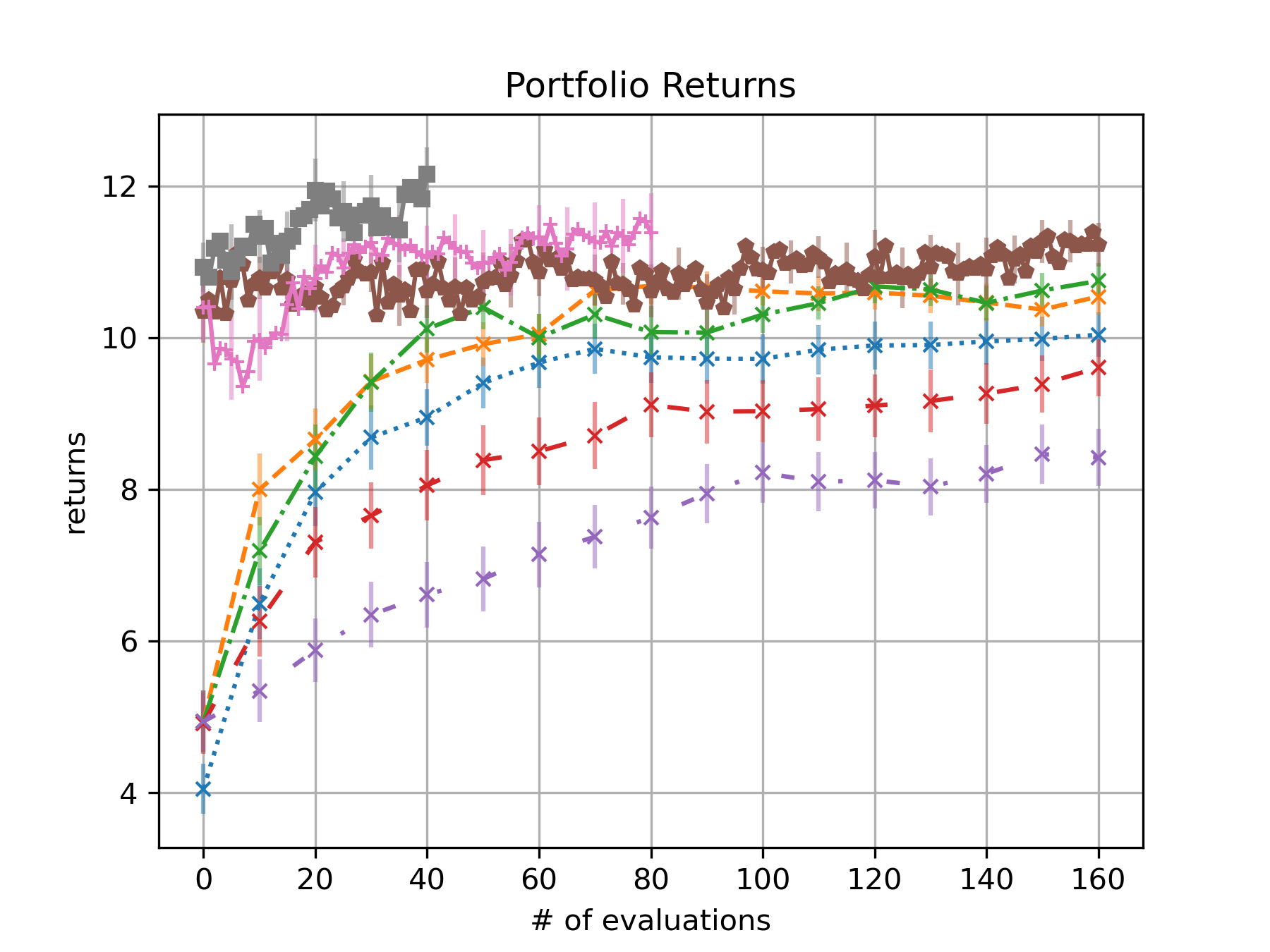} & \includegraphics[width=\linewidth]{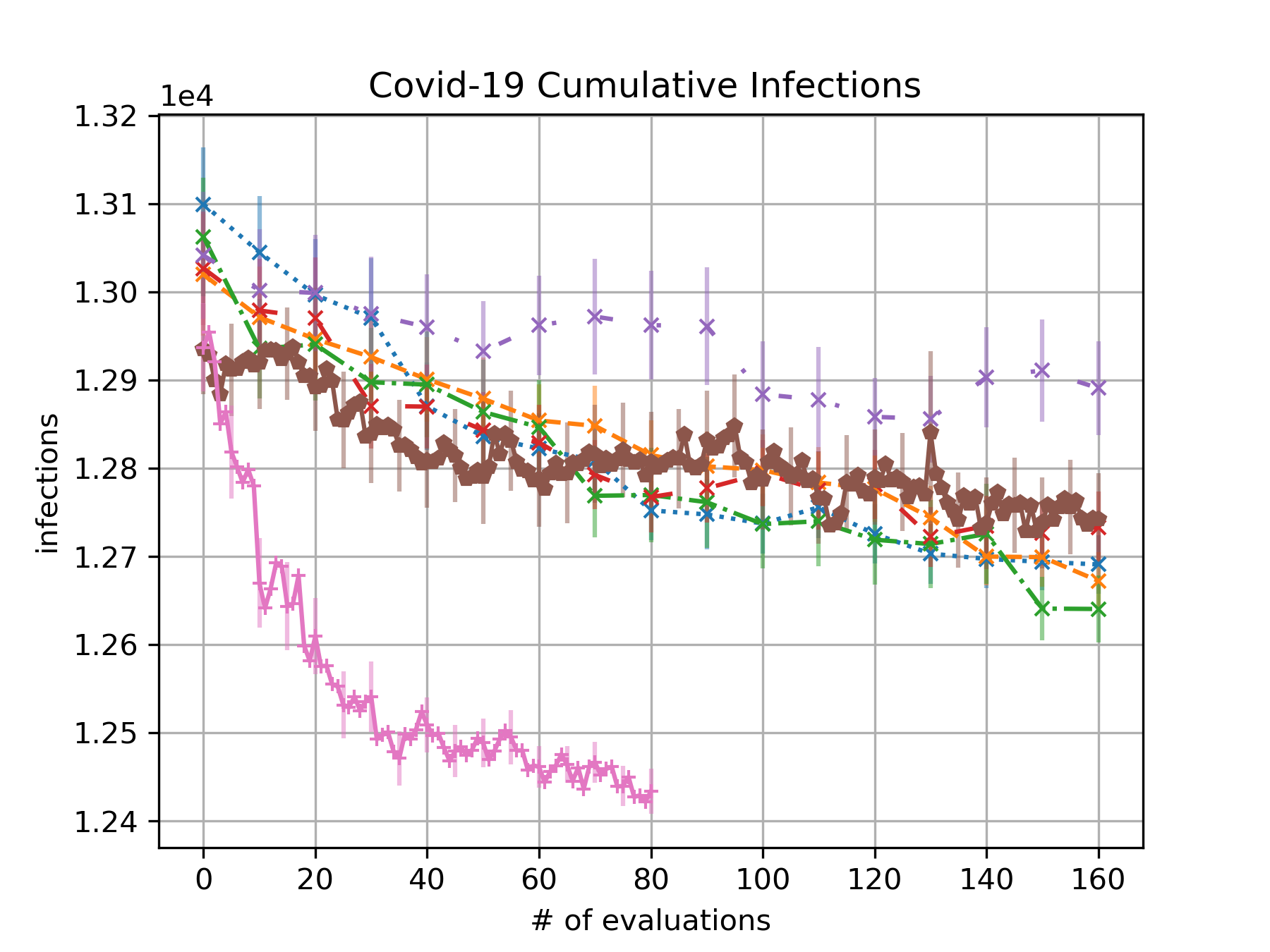} & \includegraphics[width=\linewidth]{figures/legend.png}  \\
    \end{tabular}

    \caption{Top: The log optimality gap in Branin Williams with VaR (left) with CVaR (middle), and $f_6(x_c, x_e)$ (right). Bottom: The returns on Portfolio problem (left), the cumulative number of infections in the COVID-19 problem (middle) and the legend (right). The plots are plotted against the number of $F(x, w)$ evaluations.}
    \label{fig:exp-ns-results}
\end{figure}

In the main body of the paper, we present the results using a moving average window of $3$ iterations, to smooth out the oscillations and make the plots easier to read. In Figure \ref{fig:exp-ns-results}, we present the non-smoothed plots 
to complement the moving averages presented in the paper.

% to address any questions the smoothing may raise \ezedit{what does "this" refer to? - what are the questions? Change the phrase "non-smoothto complement ..."}. 
The comparatively larger variance of our algorithms is attributed to: (i) 
our methods reporting a new solution $\sim$10x more frequently (after a single evaluation of $F(x,w)$, rather than an evaluation at many $w$), leading to more points in the plot; (ii) our methods having fewer replications than the benchmark algorithms.

\subsection{Results comparing algorithm run times}
\begin{figure}[ht]
    \centering
      \includegraphics[width=0.4\linewidth]{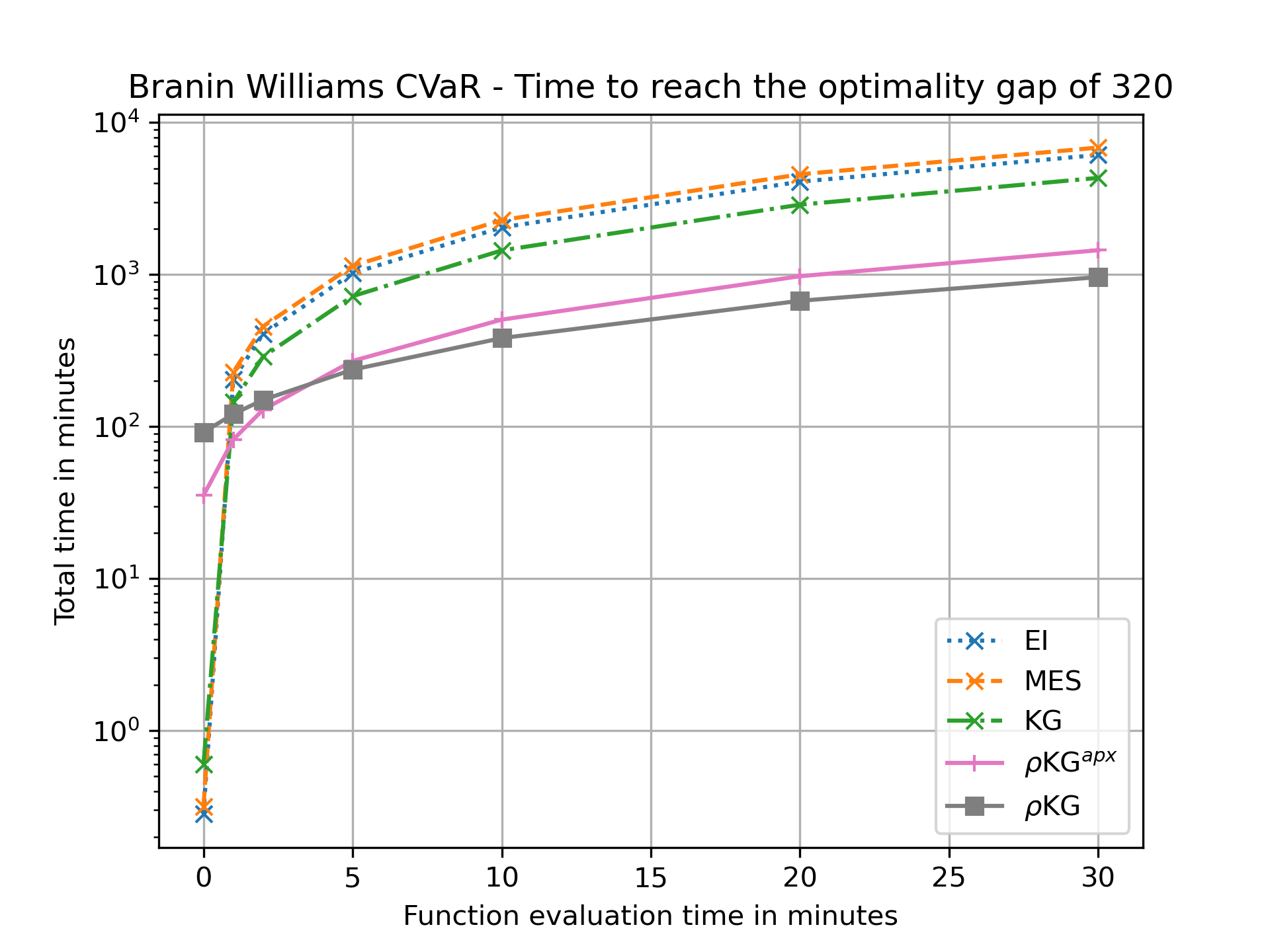} \includegraphics[width=0.4\linewidth]{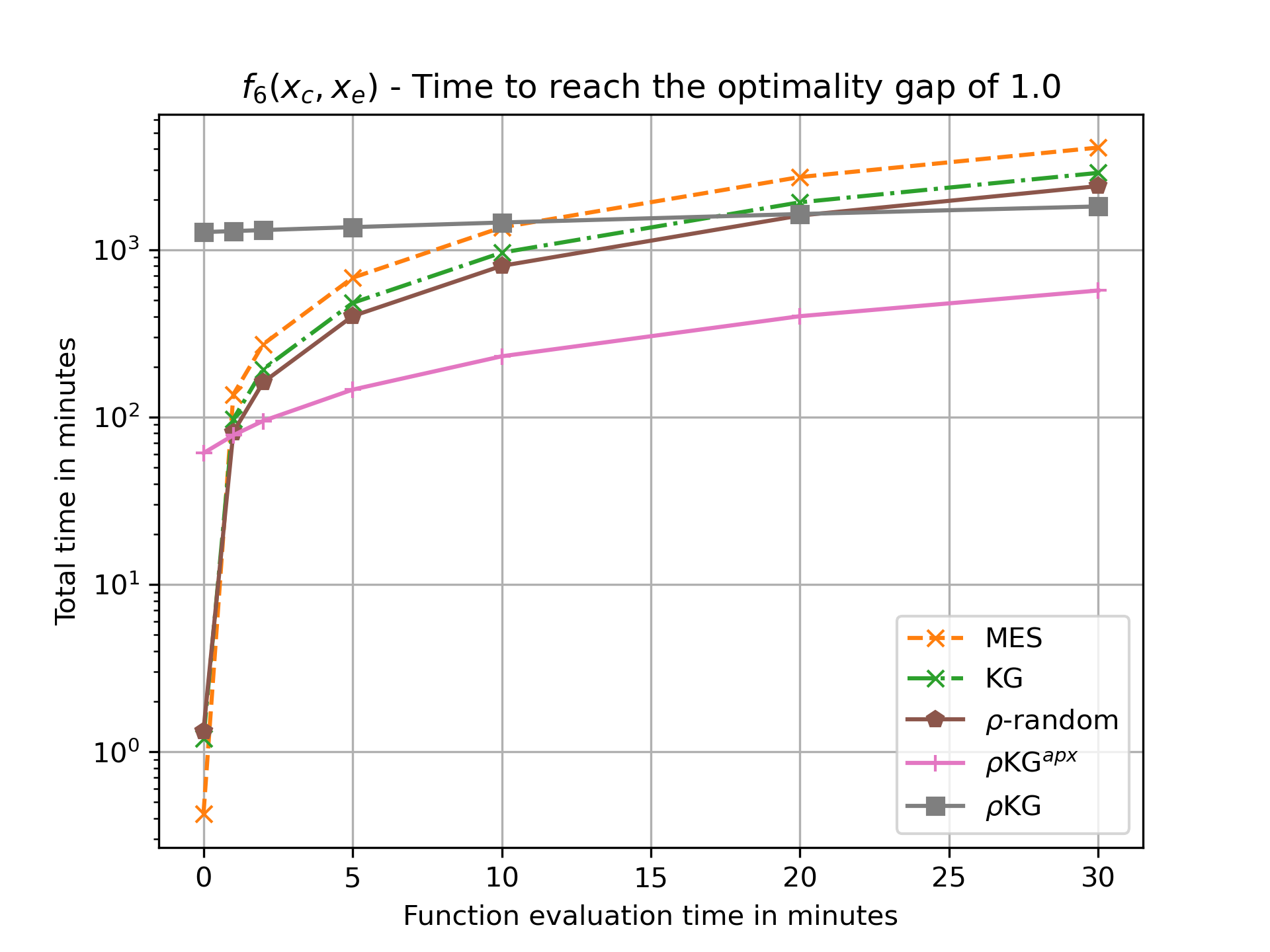}
    \caption{Left: Time to reach an optimality gap of 320 on the Branin Williams problem with CVaR objective. Right: Time to reach an optimality gap of 1 on the $f_6(x_c, x_e)$ function. The plots show the total run time vs the time per a single evaluation of the function $F(x, w)$.}
    \label{fig:time-to-benchmark}
\end{figure}
In Figure \ref{fig:time-to-benchmark}, we present plots that show the total time it takes for the algorithms to reach a given optimality gap threshold, as a function of the cost of a single evaluation of the function $F(x, w)$. 
The plots make clear the advantage of using our acquisition functions, instead of the benchmarks studied in the paper, even with only moderately expensive function evaluations. 
The Branin Williams and $f_6(x_c, x_e)$ were chosen since the results for these experiments allow for a meaningful comparison; and that they are representative as, roughly speaking, the least and the most computationally intensive in terms of acquisition function optimization, respectively, of all the numerical examples presented in the paper. It is seen that in $f_6(x_c, x_e)$, the $\acqf$ algorithm is quite expensive to optimize, and is not a good contender unless the function evaluations are expensive enough. However, for both examples, $\acqfapx$ proves to be a better option than all the benchmarks, even when the function evaluations take only a few minutes.

\section{Proofs of theoretical results} \label{sec-proofs}

%We start with a simple result showing that the value of information defined by $\acqf$ is non-negative.
%\begin{align*}
    %\E_n[ \min_x \E_n [\rho[F(x,W)]] ] &\geq \E_n[ \E_n [ \min_x \E_{n+1} [ \rho[F(x, W)] ]  \mid (x_{n+1}, w_{n+1}) ]  ] \\
    %&= \E_n [ \min_x \E_{n+1} [ \rho[F(x, W)] ] \mid (x_{n+1}, w_{n+1}) ] \\
    %& \implies \acqf(x, w) \geq 0 \ \forall (x, w)
%\end{align*}

This section is dedicated to showing that the derivative estimators we propose in this paper are asymptotically unbiased and consistent. Here, we only present the proofs for the case of $d^\W = 1$.

The derivative of a GP is again a GP (cf. \cite{wu2017gradient} and section 9.4 of \cite{Rasmussen2006GP}). If we model $F(x)$ as a GP with mean and covariance functions $\mu, \Sigma$, then $F$ and $\nabla F$ follow a multi output GP with mean and covariance functions given by \cite{wu2017gradient}
\begin{equation*}
    \widetilde{\mu}(x) = (\mu(x), \nabla \mu(x))^{\top} \text{ and } \widetilde{\Sigma}(x, x') = \left( \begin{array}{cc}
        \Sigma(x, x') & \nabla \Sigma(x, x') \\
        \nabla \Sigma(x', x)^{\top} & \nabla^2 \Sigma(x, x')
    \end{array} \right);
\end{equation*}
where $\nabla \Sigma$ and $\nabla^2 \Sigma$ are the Jacobian and Hessian of the covariance function. We note that, although this result is found in the literature without any assumptions on the GP, it in fact requires the GP to possess differentiable sample paths. For example, the Brownian motion is a well known example of a GP that has nowhere differentiable sample paths, thus violates this framework. To this end, we refer the reader to \cite{Adler07RandomFields} for precise conditions on the differentiability of the GP sample paths, and {\it assume} that these conditions are satisfied.

\begin{remark}
    Before discussing the assumptions below, we note that the assumptions listed here are more restrictive than needed. The main results we build on require certain conditions on the distribution of the function in a neighborhood of its VaR. However, when the function is drawn from a GP, it is impossible to know where this neighborhood, which would be different for each draw, might be, and thus it is impossible to impose assumptions on such an unknown and ever changing neighborhood. The main purpose of the results presented here is to show that there indeed exists a set of conditions under which our claims hold. However, we are confident that our algorithms are applicable in much broader settings, even when no such assumption can be verified.
\end{remark}

We {\it assume} that the kernel $\widetilde{\Sigma}$ of the joint GP of the function and its derivative is strictly positive definite; and the random environmental variable $W$ admits a strictly positive and continuous density $p(w)$ over its domain. (Note that this assumption is easily satisfied by adding a small Gaussian noise to $W$.)
Observe that when the kernel is strictly positive definite, the sample paths of the GP are non-zero w.p.1., i.e. the equation $F(x, w) = 0$ has at most countably many solutions, where $F(\cdot, \cdot)$ denotes a sample drawn from the GP.

\begin{lemma}\label{lemma-gradient-condition}
    Under conditions outlined above, for any $t$ such that $P_W(F(x, W) \leq t) \in (0,1)$, the CDF $P_W(F(x, W) \leq t)$ is continuously differentiable w.r.t. both $x$ and $t$; and the density $\partial_t P_W(F(x, W) \leq t)$ is strictly positive for each $x$.
\end{lemma}
\begin{proof}
    Since there are at most countably many points where the derivative is zero, we can construct countably many distinct intervals $\W_i, i \in I$ where $F(x, w)$ is monotone (in $w$) in each $\W_i$, and $\W_i$ are such that $P_W(\cup_{i \in I} \W_i) = 1$ and $P_W(\W_i) > 0, i \in I$. Then,
    \begin{align*}
        \partial_x P_W(F(x, W) \leq t) &= \partial_x \int_{\{w: F(x, w) \leq t\}} p(w) dw \\
        &= \partial_x \sum_{i \in I} P_W(\W_i) \int_{\{w \in \W_i: F(x, w) \leq t\}} \frac{p(w)}{P_W(\W_i)} dw \\
        &= \sum_{i \in I} \partial_x P_W(\W_i) \int_{\{w \in \W_i: F(x, w) \leq t\}} p(w) dw
        &= (*)
    \end{align*}
    where the interchange of derivative and summation is justified as the sum is bounded. From here on, we ignore the $\W_i$ that do not produce any root to the equation $F(x, w) = t$, as the corresponding derivatives are zero. Since $F(x, w)$ is continuously differentiable and monotone in $w$ in each $\W_i$, by the inverse function theorem \cite{Rudin1976Analysis}, there exists a continuously differentiable inverse function $F_i^{-1}(t'; x)$ mapping $t'$ to $w \in \W_i$ for a given $x$. For the remaining $\W_i$, define the sets $I^+$ and $I^-$ such that, $I^+$ is the set of $\W_i$ that has $F^{-1}_i(t; x)$ as the upper bound of the integral and $I^-$ as the ones with the lower bound. For $i \in I^+$, let $w_i^-$ be the lower bound of the integral, and similarly $w_i^+$ as the upper bound for $I^-$.  Then,
    \begin{align*}
        (*) &= \sum_{i \in I^+} \partial_x \int_{w_i^-}^{F_i^{-1}(t; x)} p(w) dw + \sum_{i \in I^-} \int_{F^{-1}(t; x)}^{w_i^+} p(w) dw \\
        &= \sum_{i \in I^+} p(w) \partial_x F_i^{-1}(t; x) - \sum_{i \in I^-} - p(w) \partial_x F_i^{-1}(t; x) \\
        &= \sum_{i \in I^+ \cup I^-} p(w) \partial_x F_i^{-1}(t; x)
    \end{align*}
    where the result follows by the Leibniz rule. Applying the same steps with $\partial_t$, we get
    \begin{equation*}
        \partial_t P_W(F(x, W) \leq t) = \sum_{i \in I^+ \cup I^-} p(w) \partial_t F_i^{-1}(t; x).
    \end{equation*}
    Since $p(w)$ is continuous and $F_i^{-1}(t; x)$ is continuously differentiable, $P_W(F(x, W) \leq t)$ is continuously differentiable w.r.t. both $x$ and $t$. Since $p(w)$ is strictly positive and the derivative is non-zero, it follows that $\partial_t P_W(F(x, W) \leq t)$ is strictly positive, thus completing the proof.
\end{proof}

\begin{proposition} \label{proposition-inner-gradient}
    Under the conditions outlined above, for $\alpha \in (0, 1)$,
    \begin{equation*}
       \frac{1}{M} \sum_{j=1}^M \nabla_x \widehat{v}^j (x) \xrightarrow{p} \nabla_x \E_{n+1} [\text{VaR}_{\alpha}[F(x, W)]]
    \end{equation*}
    as $M, L \rightarrow \infty$, where $\xrightarrow{p}$ denotes convergence in probability. Moreover, the gradient $\nabla_x \E_{n+1} [\text{VaR}_{\alpha}[F(x, w)]]$ is continuous.
\end{proposition}
\begin{proof}
    Let $F(x, w)(\omega)$ denote a sample path of the GP.
    As a sample path of the GP, $F(x, w)(\omega)$ is continuously differentiable w.r.t. $w$ w.p.1. By Lemma \ref{lemma-gradient-condition}, the CDF $P_W(F(x, W) \leq t)$ is continuously differentiable w.r.t. both $x$ and $t$, and the density $\partial_t P_W(F(x, W) \leq t)$ is strictly positive. Then, by Lemma 2 of \cite{Jiang2015Quantile},
    \begin{equation}\label{eq-limit-var-gradient}
        \lim_{L \rightarrow \infty} \E[\nabla_x \widehat{v}(x)] = \nabla_x \text{VaR}_{\alpha}[F(x, W)].
    \end{equation}
    
    Since the gradient is also a GP defined in a compact space, $\nabla_x F(x, w)(\omega)$ is bounded w.p.1., so is $\nabla_x \text{VaR}_{\alpha}[F(x, W)]$. Then, by proposition 1 of \cite{Broadie1996IPA},
    \begin{equation}\label{eq-var-gradient-ipa-interchange}
        \nabla_x \E_{n+1} [\text{VaR}_{\alpha}[F(x, W)]] = \E_{n+1}[ \nabla_x \text{VaR}_{\alpha}[F(x, W)] ]. 
    \end{equation}
    $F(x, w)$ is bounded by the same argument as $\nabla_x F(x, w)$, thus has finite second moment. Therefore, the main result follows by an application of Chebyshev's inequality. The continuity of $\nabla_x \E_{n+1} [\text{VaR}_{\alpha}[F(x, W)]]$ follows by Theorem 1 of \cite{Jiang2015Quantile} and the fact that both gradients are continuous as shown in Lemma \ref{lemma-gradient-condition}.
\end{proof}

Before going into the main result, we discuss the mapping from the candidate point $(x, w)$ to the sample $\widehat{F}^{ij}(\cdot, \cdot)$, the $j$-th sample path drawn from the $i$-th fantasy GP model, $\gp_f^i$. Let's start with the mapping from the candidate to the mean and covariance functions of $\gp_f^i$. Let $Z_i$ be (a fixed realization of) the standard normal random variable used to generate the fantasy. The mean and covariance functions of $\gp_f^i$ are given by:
\begin{align*}
    \mu_f^i(x', w') &= \mu_n(x', w') + \frac{\Sigma_n(x', w', x, w)}{\sqrt{\Sigma_n(x, w, x, w) + \sigma^2(x, w)}} Z_i, \\
    \Sigma_f^i(x', w', x'', w'') &= \Sigma_n(x', w', x'', w'') - \frac{\Sigma_n(x', w', x, w) \Sigma_n(x, w, x'', w'')}{\Sigma_n(x, w, x, w) + \sigma^2(x, w)},
\end{align*}
where $\sigma^2(x, w)$ is the known observation noise level (see \cite{Wu2016ParallelKG} for more details). {\it Assuming} that the prior mean and covariance functions are continuously differentiable, $\mu_f^i$ and $\Sigma_f^i$ are continuously differentiable w.r.t. the candidate $(x, w)$. 

For a given $x^i$ and finite set $\widetilde{W} = \{w_{1:L}\}$, and a (realization of) $L$ - dimensional standard normal random vector $Z_L^{ij}$, the $j$-th sample path of $i$-th fantasy model is generated as
\begin{equation}\label{eq-sample-from-gp}
    \widehat{F}^{ij}(x^i, w_{1:L}) = \mu_f^i(x^i, w_{1:L}) + C(x^i, w_{1:L}) Z_L^{ij},
\end{equation}
where $C(x^i, w_{1:L})$ is the Cholesky factor of $L \times L$ covariance matrix $\Sigma_f^i(x^i, w_{1:L}, x^i, w_{1:L})$. Since $\Sigma_f^i$ is continuously differentiable w.r.t. the candidate, so is $C(x^i, w_{1:L})$, making the sample $\widehat{F}^{ij}(\cdot, \cdot)$ continuously differentiable w.r.t. the candidate $(x, w)$.

The discussion above establishes continuous differentiability of the sample $\widehat{F}^{ij}(\cdot, \cdot)$ w.r.t. the candidate $(x, w)$ only for a finite collection of points $(x^i, w_{1:L})$. However, to be able to use the analysis of Proposition \ref{proposition-inner-gradient}, we need continuous differentiability of $\widehat{F}^{ij}(x, w), w \in W$ which, with $W$ being a continuous random variable, is an infinite dimensional random variable. The concept of an infinite dimensional standard Gaussian random variable, or a standard Gaussian random function, is not well defined, and we can no longer use equation (\ref{eq-sample-from-gp}). In the proposition below, we {\it assume} that $F(\cdot, \cdot)$ is indeed differentiable w.r.t. the candidate $(x, w)$ and hope that this discussion justifies the assumption. Below, we use $u := (x, w)$ as the candidate to simplify the notation.

\begin{proposition}\label{proposition-outer-gradient}
    Under the conditions outlined above, and assuming that the solution to the inner problem $x^*$ is unique w.p.1., 
    \begin{equation}
        \frac{1}{K} \sum_{i=1}^K - \frac{1}{M} \sum_{j=1}^M \nabla_u \widehat{v}^j(x^*_i) \xrightarrow{p} \nabla_u \E_n [ - \min_x \E_{n+1} [\text{VaR}_{\alpha}[F(x, w)]] \mid u ]
    \end{equation}
    as $K, L, M \rightarrow \infty$ 
    % \ezedit{is this simultaneous limit, or iterate limit? from the proof, it seems like iterate limit, i.e., lim_L lim_{K,M} (not sure if K and M are also iterative)}.
\end{proposition}
\begin{proof}
    Following the proof of proposition \ref{proposition-inner-gradient} with $\nabla_u$ replacing $\nabla_x$, we get
    \begin{equation*}
        \lim_{L \rightarrow \infty} \E_{n+1}[ \E[\nabla_u \widehat{v}(x)] ] = \nabla_u \E_{n+1} [\text{VaR}_{\alpha}[F(x, W)]].
    \end{equation*}
    By the assumption that $x^*$ is unique w.p.1. and the continuity of $\nabla_u E_{n+1} [\text{VaR}_{\alpha}[F(x, W)]]$ by proposition \ref{proposition-inner-gradient}, we can apply the envelope theorem (corollary 4, \cite{milgrom2002envelope}) to get
    \begin{equation*}
        \lim_{L \rightarrow \infty} \E_{n+1}[ \E[\nabla_u \widehat{v}(x^*)] ] = \nabla_u \E_{n+1} [\text{VaR}_{\alpha}[F(x^*, W)]].
    \end{equation*}
    Again by the arguments in the proof of proposition \ref{proposition-inner-gradient}, the gradient is continuous and bounded w.p.1., thus applying proposition 1 of \cite{Broadie1996IPA}, we get
    \begin{equation}\label{eq-outer-unbiased}
        \lim_{L \rightarrow \infty} \E_n[ \E_{n+1}[ \E[\nabla_u \widehat{v}(x^*)] ] \mid u] = \nabla_u \E_n[ \E_{n+1} [\text{VaR}_{\alpha}[F(x^*, W)]] \mid u].
    \end{equation}
    Since the second moment of the estimators are bounded (see the proof of proposition \ref{proposition-inner-gradient}), the result follows by Chebyshev's inequality.
\end{proof}

\begin{proposition}
    The results presented in Propositions \ref{proposition-inner-gradient} \& \ref{proposition-outer-gradient} extend to the risk measure CVaR, i.e. 
    \begin{equation*}
       \frac{1}{M} \sum_{j=1}^M \nabla_x \widehat{c}^j (x) \xrightarrow{p} \nabla_x \E_{n+1} [\text{CVaR}_{\alpha}[F(x, W)]];
    \end{equation*}
    and 
    \begin{equation} 
        \frac{1}{K} \sum_{i=1}^K - \frac{1}{M} \sum_{j=1}^M \nabla_u \widehat{c}^j(x^*_i) \xrightarrow{p} \nabla_u \E_n [ - \min_x \E_{n+1} [\text{CVaR}_{\alpha}[F(x, w)]] \mid u ]
    \end{equation}
    as $K, L, M \rightarrow \infty$.
    % \ezedit{same problem as above}.
\end{proposition}
\begin{proof}
    We have established in the proof of Proposition \ref{proposition-inner-gradient} that $\nabla F(x, w)$ is bounded, and that $\text{VaR}_{\alpha}[F(x, W)]$ is continuously differentiable. Combining these with the assumption that $W$ is a continuous random variable, we can apply Theorem 3.1 of \cite{Hong2009CVaR} to get
    \begin{equation*}
        \nabla \text{CVaR}_{\alpha}[F(x, W)] = \E[\nabla F(x, W) \mid F(x, W) \geq \text{VaR}_{\alpha}[F(x, W)]].
    \end{equation*}
    Since $\nabla F(x, W)$ is continuous and bounded, it follows by bounded convergence theorem \cite{Durrett2010PTE} that $\text{CVaR}_{\alpha}[F(x, W)]$ is continuously differentiable. The remaining arguments are identical to those of Propositions \ref{proposition-inner-gradient} \& \ref{proposition-outer-gradient}.
\end{proof}

We conclude this discussion by noting that, while the proposition in the main paper claims both asymptotic unbiasedness and consistency, the propositions proved here only establish the consistency of the estimators, and the asymptotic unbiasedness result is hidden within the proof argument. In the proof of Proposition \ref{proposition-inner-gradient}, combining the equations (\ref{eq-limit-var-gradient}) and (\ref{eq-var-gradient-ipa-interchange}) gives us the asymptotic unbiasedness of the gradient estimators for the inner problem, where the interchange of limit and expectation is justified by the dominated convergence theorem \cite{Durrett2010PTE}. Similarly, equation (\ref{eq-outer-unbiased}) shows that the estimators for the candidate optimization of $\acqf$ are asymptotically unbiased.

\small

\bibliographystyle{ieeetr} 
\bibliography{ref.bib}